\documentclass[lettersize,journal]{IEEEtran}
\usepackage{amsmath,amssymb,amsfonts}
\usepackage{algorithmic}
\usepackage{array}
\usepackage{textcomp}
\usepackage{stfloats}
\usepackage{url}
\usepackage{verbatim}
\usepackage{graphicx}
\usepackage{cite}

\usepackage{tabularx}
\usepackage{wrapfig}
\usepackage{makecell}
\usepackage{booktabs}
\usepackage{comment}
\usepackage{subfigure}
\usepackage{booktabs}
\usepackage[ruled,vlined,linesnumbered]{algorithm2e}
\usepackage{multirow}
\usepackage{hyperref}
\usepackage[nameinlink]{cleveref}
\Crefname{figure}{{Fig.}}{{Figs.}}
\Crefname{equation}{{}}{{}}
\Crefname{table}{{Tab.}}{{Tabs.}}

\newcommand{\etal}{\emph{et al.}\xspace}
\newcommand{\eg}{\emph{e.g.,}\xspace}
\newcommand{\Eg}{\emph{E.g.,}\xspace}

\hypersetup{
  colorlinks=true,
  citecolor=blue,
  linkcolor=blue,
  urlcolor=green}

\begin{document}

\title{A Bionic Data-driven Approach for \\Long-distance Underwater Navigation with Anomaly Resistance}

\author{Songnan Yang, Xiaohui Zhang, Shiliang Zhang,~\IEEEmembership{Member,~IEEE}, Xuehui Ma, Wenqi Bai, Yushuai Li,~\IEEEmembership{Member,~IEEE}, Tingwen Huang,~\IEEEmembership{Fellow,~IEEE}
\thanks{This work was supported in part by the National Major Scientific Instrument Development Project of China under Grant 62127809; in part by the National Natural Science Foundation of China under Grant 62073258; and in part by the Basic Research in Natural Science and Enterprise Joint Foundation of Shaanxi Province under Grant 2021JLM-58. (Corresponding author: Xiaohui Zhang.) }
\thanks{Songnan Yang, Xiaohui Zhang, and Wenqi Bai are with the School of Automation and Information Engineering, Xi’an University of Technology, Shaanxi 710048, China, and are also with the Institute of Advanced Navigation and Electromagnetics, Xi’an University of Technology, Shaanxi 710048, China (e-mail: yang.son.nan@gmail.com, xhzhang@xaut.edu.cn, bayouenqy@outlook.com).}
\thanks{Xuehui Ma is with the School of Automation and Information Engineering, Xi’an University of Technology, Shaanxi 710048, China (e-mail: xuehui.yx@gmail.com).}
\thanks{Shiliang Zhang is with the Department of Informatics, University of Oslo, 0316 Oslo, Norway (e-mail: shilianz@ifi.uio.no).}
\thanks{Yushuai Li is with the Department of Computer Science, Aalborg University, 9220 Aalborg, Denmark (e-mail: yushuaili@ieee.org).}
\thanks{Tingwen Huang is with Texas A\&M University at Qatar, Doha 23874, Qatar (e-mail: tingwen.huang@qatar.tamu.edu).}
}



\maketitle

\begin{abstract}
Various animals exhibit accurate navigation using environment cues. The Earth's magnetic field has been proved a reliable information source in long-distance fauna migration. Inspired by animal navigation, this work proposes a bionic and data-driven approach for long-distance underwater navigation. The proposed approach uses measured geomagnetic data for the navigation, and requires no GPS systems or geographical maps. Particularly, we construct and train a Temporal Attention-based Long Short-Term Memory (TA-LSTM) network to predict the heading angle during the navigation. To mitigate the impact of geomagnetic anomalies, we develop the mechanism to detect and quantify the anomalies based on Maximum Likelihood Estimation. We integrate the developed mechanism with the TA-LSTM, and calibrate the predicted heading angles to gain resistance against geomagnetic anomalies. Using the retrieved data from the WMM model, we conduct numerical simulations with diversified navigation conditions to test our approach. The simulation results demonstrate a resilience navigation against geomagnetic anomalies by our approach, along with precision and stability of the underwater navigation in single and multiple destination missions.
\end{abstract}


\section{Introduction}
\label{sec:introduction}
\IEEEPARstart{U}{nderwater} navigation empowers exploration and operation in the sea, and has drawn attention in a wide range of applications such as resource survey, rescue, mapping, and underwater maintenance~\cite{Zhang2021a}. The navigation information is critical for marine carriers to approach the destination, conduct underwater tasks, and return their base safely and timely. Nevertheless, due to the underwater environment where radio signals attenuate rapidly, navigation techniques used on the land or in the air, \eg, the Global Navigation Satellite System (GNSS), do not work any longer~\cite{MELO2017250}.

Numerous alternative solutions are available for underwater navigation, mainly the Inertial Navigation System (INS), acoustic navigation, terrain matching navigation, gravity matching navigation, and geomagnetic matching navigation. However, those solutions cannot fully address the challenges in underwater navigation, especially for long-distance missions. \Eg using INS in underwater navigation is a passive method that does not rely on external information, yet INS induces positioning deviation due to the error drift of inertial devices. This error drift accumulates over time and distance, impeding the accuracy for long-distance navigation~\cite{DBLP:journals/tie/ShenZGCCTLL21}. Acoustic navigation can provide precise real-time location, but it requires the deployment of signal beacons ahead of the mission~\cite{DBLP:journals/ral/ChoiBSC22}, which is impractical in unknown environment in long-range navigation. Geophysical matching navigation approaches, including terrain matching~\cite{MELO2017250}, gravity matching~\cite{DBLP:journals/tim/WangHWWZ22}, and geomagnetic matching~\cite{DBLP:journals/inffus/LiZZLNE17}, can provide strong autonomy for underwater navigation and they can work under diverse terrain and time conditions. Such approaches refers to the navigation based on comparing the similarity between the navigation trajectory and a reference map~\cite{DBLP:journals/tgrs/ChenLZLCPHX23}. The terrain contour matching (TERCOM) algorithm~\cite{10.1117/12.959127} receives continuous navigation corrections throughout the mission, and it can reserve the positioning accuracy in cases of large initial position error and altimeter measurement noise. Gravity matching determines the carrier location by comparing the measured gravity field with a reference gravity map. A gravity matching approach by Wang \etal~\cite{7387748} looks into the correlation between adjacent locations provided by INS, and improves the navigation robustness in gravity anomaly areas. Magnetic contour matching (MAGCOM)~\cite{DBLP:journals/inffus/LiZZLNE17} is often used in conjunction with INS~\cite{DBLP:journals/lgrs/ChenZPCWWL18}, because its initial heading error can lead to large position errors thus requires calibration. Overall, the geophysical matching navigation approaches largely depend on the reference map and the accuracy and completeness of the pre-stored map data, which is nontrivial for long-distance navigation and hard to guarantee for enormous reasons~\cite{Zhang2021}.

While we observe a scarcity of pre-stored information or map data for long-distance navigation, such scarcity does not hold for animal navigation. Numerous organisms are known to migrate thousands of kilometers and then return their base accurately~\cite{geva2015spatial}. It appears that such animals do not have large-scale geographic map data stored in their brains~\cite{Zhang2021}, instead, they depend on real-time information perceived for long-distance navigation.

Signals in the environment like the distribution of water temperature, currents, and pressure provides little information for animal navigation~\cite{DBLP:journals/tgrs/ZhaoHCHLR14}. Increasing evidence reveals that animals can use the geomagnetic signals from the Earth to orient and find their position~\cite{putman2020animal}. Magnetic displacement experiments~\cite{gould2014animal} have illustrated various taxa, including crustaceans, fish, amphibians, reptiles, and birds~\cite{boles2003true,naisbett2017magnetic,fischer2001evidence,lohmann2004geomagnetic,pakhomov2018magnetic}, can derive vital navigational information from geomagnetic signals. 

Inspired by animal navigation, bionic navigation has drawn attention and navigation approaches using geomagnetic information have gained momentum. Qi \etal~\cite{DBLP:journals/tvt/QiCAWZXYLR18} developed an Extended Kalman Filer (EKF) based algorithm to predict heading angles for the navigation. They simulated the magnetic dipole of the Earth and used the magnetic gradient for the heading angle prediction. However, variations, irregularities, and anomalies can exist in the real geomagnetic field~\cite{DBLP:journals/tgrs/ZhaoHCHLR14}, which can reduce the accuracy of heading angle prediction based on the simulated geomagnetic field. Taylor \etal~\cite{taylor2021long} proposed a geomagnetic navigation method using the geomagnetic inclination for long-distance navigation, and their simulation results show the capacity of autonomous transequatorial navigation in both present-day and reversed magnetic fields. Nevertheless, the performance of their approach significantly decreases when there is magnetic noise. Therefore, it is essential to mitigate the impact of geomagnetic irregularities and anomalies in geomagnetic navigation to fit the real world scenarios~\cite{KIM2023112706,DBLP:journals/taes/Canciani22}.

Several studies have attempted to address the anomaly issues in geomagnetic navigation. Zhang \etal~\cite{DBLP:journals/tim/ZhangZSWZ21} proposed a geomagnetic gradient-assisted evolutionary algorithm for long-range underwater navigation. Their approach constructs an evolutionary sample space, and they use the geomagnetic gradient and the constructed space to optimize the navigation in geomagnetic anomaly areas. However, their approach involves ineffective searching computations in optimizing their objective function, implying the need for a more thorough method beyond the monotonic decline of the objective function. Zhang \etal~\cite{DBLP:journals/tcyb/ZhangLLY21} proposed a geomagnetic navigation method using model predictive control (MPC). Their approach requires no prior knowledge of the geographical location or geomagnetic map. This method demonstrates the potential of using magnetic declination and inclination in long-distance underwater navigation. Their simulation results indicate robustness of the navigation in mitigating the impact of geomagnetic anomalies. Nevertheless, they use fixed $Q$ and $R$ parameters in heading angle prediction for the navigation, which hinders the adaptability and accuracy of their approach, especially in geomagnetic anomaly areas where the magnetic signals vary dramatically. Qi \etal~\cite{DBLP:journals/tgrs/QiXXLR23} developed a 2-D approach  for long-distance underwater navigation. They achieve the navigation by heading angle prediction using geomagnetic gradients. They provided theoretical analysis and simulations for the robustness of their approach under interference and measurement errors. Yet they did not consider the situation where the carrier falls into a local area with unknown geomagnetic anomalies. The geomagnetic anomaly issue remains challenging in that (i) geomagnetic anomalies can vary over time and is hard to predict, making it impractical to create a reference map for long-distance geomagnetic navigation (ii) geomagnetic anomaly is almost inevitable in long-distance missions, leading to obstacles for the navigation based on geomagnetic information, and (iii) geomagnetic anomalies render the calculation of geomagnetic gradient useless, thus deteriorating the heading angle prediction based on the gradients and resulting in unnecessary travel or even failed geomagnetic navigation.

In this work, we propose an approach for long-distance navigation with anomaly resistance. Our approach is data-driven that uses measured geomagnetic data to predict heading angle for the navigation, and requires no geographical or geomagnetic map for reference. Particularly, we encode the geomagnetic information along consecutive locations during the navigation as time-series, and we predict the heading angle using the time-series and intermediate calculated results. For the heading angle prediction, we construct and train a Temporal Attention based Long Short-Term Memory (TA-LSTM) network based on the encode time-series. The temporal attention mechanism in the TA-LSTM enables the training of the TA-LSTM focus on the most pertinent geomagnetic elements in the time-series, which are the ones that contribute to the heading angle prediction. In this way, the TA-LSTM training process limits the impact of geomagnetic anomalies in local areas. To further resist the anomalies, we develop a mechanism to detect and quantify geomagnetic anomalies based on Maximum Likelihood Estimation (MLE). We integrate the developed mechansim with the constructed TA-LSTM to calibrate the heading angle prediction for enhanced anomaly resistance. To the best of our knowledge, we are the first to develop a TA-LSTM based solution for long-distance underwater navigation using geomagnetic information. We summarize our contribution below.
\begin{enumerate}
    \item We propose a data-driven approach for long-distance underwater navigation. Our approach resembles animal dead-reckoning, and uses measured geomagnetic data for the navigation and requires no geographical or geomagnetic reference map.

    \item We encode consecutive geomagnetic data and intermediate calculated results into time-series, and we employ TA-LSTM to predict heading angle using the encoded time-series. The adoption of TA-LSTM reduces the impact of abnormal geomagnetic information in TA-LSTM training, thus contributing to a robust heading angle prediction in geomagnetic anomaly areas.

    \item We develop a mechanism to detect and quantify geomagnetic anomalies. We use this mechanism to calibrate the heading angle predicted by the TA-LSTM and achieve enhanced anomaly resistance.

    \item We consider consecutive geomagnetic information in heading angle prediction by the TA-LSTM. This can lead to a stable heading angle prediction compared with existing geomagnetic navigation approaches, \eg those depend their prediction on the geomagnetic gradient at a single location and are sensitive to changes in the geomagnetic field.  
\end{enumerate}

We organize our work as follows. \Cref{sec:Fundamental} presents the mathematical description of the geomagnetic field and formulates the geomagnetic navigation problem. \Cref{sec:Method} details the proposed navigation approach with anomaly resistance. \Cref{sec:simulation} conducts simulations under diverse navigation conditions, analyzes the navigation performances, and makes comparison. We conclude our work in \Cref{sec:conclusion}.

\section{Fundamental}
\label{sec:Fundamental}
\subsection{Mathematical Description of the Geomagnetic Field}
\label{sec:2.1}

The geomagnetic field (GF) is a physical field of the Earth and can be measured by a magnetometer. The GF strength $F$ at a certain location on the Earth is mainly composed of three components, $B_m$, $B_e$, and $B_d$, as shown in~\Cref{eq:1}. \({B_m}\) denotes the main magnetic field, which is produced by large scale electric currents in the liquid outer core of the earth~\cite{maus2005signature}. \({B_e}\) means the abnormal geomagnetic field generated mainly from ferrimagnetic minerals. \({B_d}\) is the geomagnetic anomaly due to changes in ionosphere and magnetosphere. While $B_d$ for a given location is associated with solar activities and can vary dramatically over time~\cite{Glatzmaier1996}, $B_m$ and $B_e$ are stable or hardly vary. 
\begin{align}
 F = {B_m} + {B_e} + {B_d}.
\label{eq:1}   
\end{align}

\begin{figure}[htbp]
\centerline{\includegraphics[width=0.8\columnwidth]{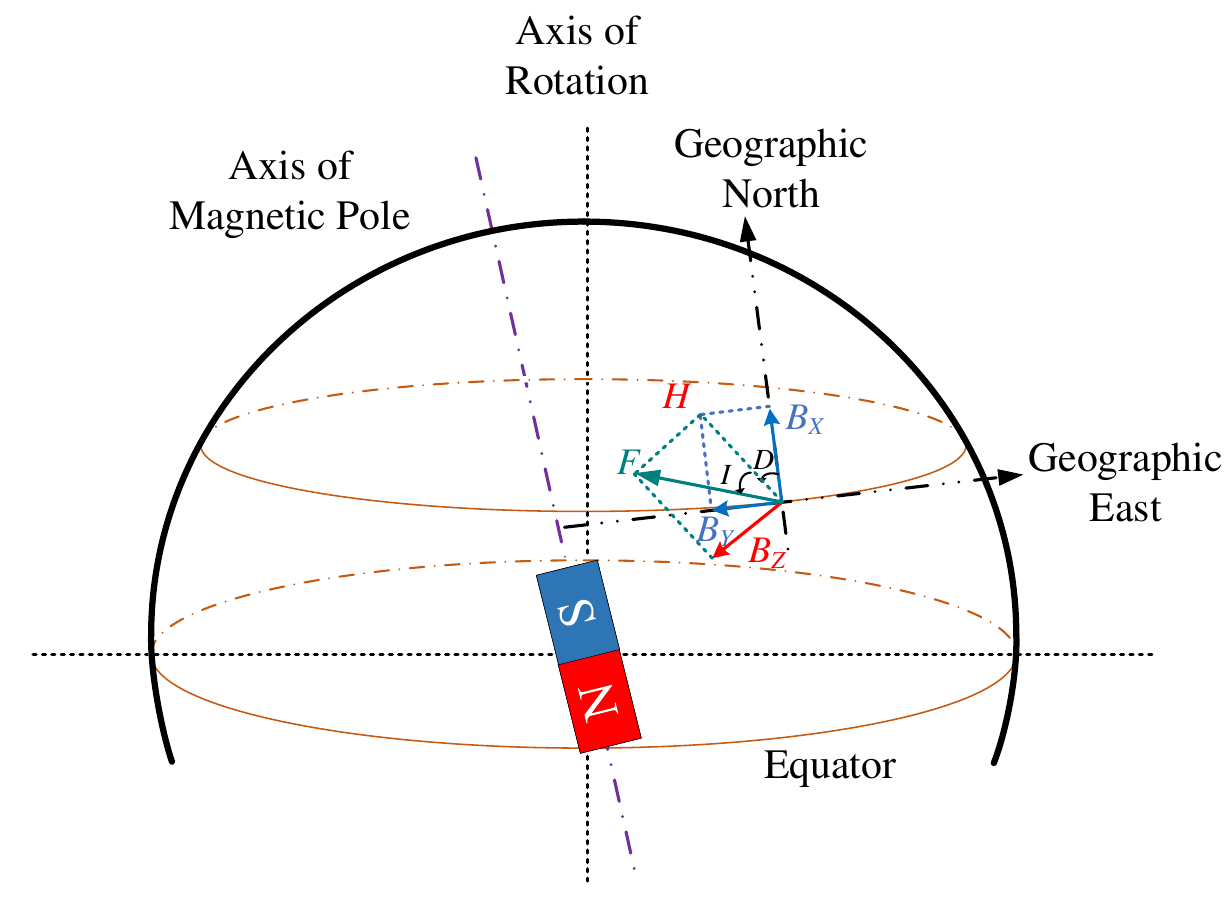}}
\caption{Main elements in the geomagnetic field.}
\label{fig:1}
\end{figure}

\Cref{fig:1} shows the main elements in GF. A mathematical description of GF can be presented by seven geomagnetic elements, as shown in \Cref{eq:2}. Parts of those elements follow two-dimension coordination, where the x-axis is along with geographical north and y-axis with geographical east. Specifically, for a given location, $F$ denotes the total strength of GF. $H$ and $B_Z$ are the projections of $F$ on the horizontal plane and in the vertical direction, respectively. $B_X$ and $B_Y$ are the projections of $H$ in the x-axis and y-axis, respectively. $I$ is the magnetic inclination presenting the angle between $F$ and the horizontal plane, and $D$ is the magnetic declination presenting the angle between $H$ and $B_X$. Note that geographical north is not the same with geomagnetic north. That is, the axis of the geomagnetic pole currently tilts at an angle of \(11^\circ \) with respect to the axis of Earth rotation or geographical north.

\begin{align}
\left\{ \begin{array}{l}{B_X} = F\cos I\cos D\\{B_Y} = F\cos I\sin D\\{B_Z} = F\sin I\\F = \sqrt {B_X^2 + B_Y^2 + B_Z^2} \\H = F\cos I\\I = \arccos (H/F)\\D = \arccos ({B_x}/H)\end{array} \right..
\label{eq:2}   
\end{align}

When describing GF using the geomagnetic vector $(B_X, B_Y, B_Z, F, H, I, D)$, theoretically, one geomagnetic vector corresponds to a unique geographical location on the Earth~\cite{Komolkin2017}, which lays a solid foundation for geomagnetic underwater navigation. Note that for the seven elements in the geomagnetic vector, four of them are derived. Therefore, it is possible to use less information in a geomagnetic vector to achieve the navigation.

\subsection{Navigation and the Geomagnetic field}
\label{sec:2.2}
Assume the actual navigation trajectory of a marine carrier, \eg an autonomous underwater vehicle (AUV), is $L$:
\begin{align}
L = \{{L_o}, {L_1}, ..., {L_{k - 1}}, {L_k}, ... \},
\label{eq:3}   
\end{align}%
where $L_k$ denotes the location at the time instance $k$. Associate those locations with a two-dimension coordinate then we get \({L_k} = {[{l_{x,k}},{l_{y,k}}]^T}\), where $l_{x,k}$ and $l_{y,k}$ are the projections of $L_k$ in the direction of geographical north and east, respectively. Then we have
\begin{align}
\left\{ \begin{array}{l}{l_{x,k + 1}} = {l_{x,k}} + \cos {\theta _k} \cdot {V_k}\\{l_{y,k + 1}} = {l_{y,k}} + \sin {\theta _k} \cdot {V_k}\end{array} \right.,
\label{eq:4}   
\end{align}
where $\theta _k$ is the heading angle of AUV that determines the movement direction, and $V_k$ denotes the AUV velocity. Below we present the trajectory in \Cref{eq:3} by geomagnetic elements shown in \Cref{eq:2}. Particularly, we associate a geomagnetic vector $S_k$ with the location $L_k$ in \Cref{eq:3} by combining two geomagnetic elements $S_k=(D_{k}, I_{k})$ , where $D_{k}$ and $I_{k}$ are the magnetic declination and inclination at location $L_k$, respectively. The relationship between adjacent declinations and inclinations is as follows
\begin{align}
\left\{ \begin{array}{l}{D_{k + 1}} = {D_k} + {g_{Dx,k}}\cos {\theta _k} + {g_{Dy,k}}\sin {\theta _k}\\{I_{k + 1}} = {I_k} + {g_{Ix,k}}\cos {\theta _k} + {g_{Iy,k}}\sin {\theta _k}\end{array} \right.,
\label{eq:5}   
\end{align}
where \({g_{Dx,k}}\) and \({g_{Dy,k}}\) are the gradients of $D$ in the direction of geographical north and east, respectively, \({g_{Ix,k}}\) and \({g_{Iy,k}}\) are the gradients of $I$ in the direction of geographical north and east~\cite{Zhang2021,Zhang2021a}, respectively, as presented below
\begin{align}
\left\{ \begin{array}{l}{g_{Dx,k + 1}} = \frac{{{D_k} - {D_{k - 1}}}}{{{l_{x,k}} - {l_{x,k - 1}}}}\\{g_{Dy,k + 1}} = \frac{{{D_{k + 1}} - {D_k}}}{{{l_{y,k + 1}} - {l_{y,k}}}}\\{g_{Ix,k + 1}} = \frac{{{I_k} - {I_{k - 1}}}}{{{l_{x,k}} - {l_{y,k - 1}}}}\\{g_{Iy,k + 1}} = \frac{{{I_{k + 1}} - {I_k}}}{{{l_{y,k + 1}} - {l_{y,k}}}}\end{array} \right..
\label{eq:6}   
\end{align}

Based on \Cref{eq:4}-\Cref{eq:6}, the heading angle \({\theta'}_k\) at the instance $k$ for the navigation can be derived as: 
\begin{align}
{\theta'}_k = \arctan (\frac{{({I_k} - {I_d}){g_{Dx,k}} - ({D_k} - {D_d}){g_{Ix,k}}}}{{({D_k} - {D_d}){g_{Iy,k}} - ({I_k} - {I_d}){g_{Dy,k}}}}).
\label{eq:7}   
\end{align}
The navigation can be achieved through the derivation of heading angle iteratively towards the destination.

The navigation process can be seen as a convergence of the geomagnetic state vector $M_{k}=(B_X, B_Y, B_Z, D, I)_k$ at the location $L_k$ toward those at the destination, represented by $M_{d}=(B_X, B_Y, B_Z, D, I)_d$. In this work, we quantify this convergence using the objective function below
\begin{align}
F_i({M_k}) = \frac{(M_{k,i}-M_{d,i})^2}{(M_{0,i}-M_{d,i})^2},
\label{eq:8}
\end{align}

\begin{align}
F({M_k}) = \frac{1}{L}\sum\limits_{i = 1}^{L} F_i({M_k}),
\label{eq:9}
\end{align}
where $L$ is the length of $M_k$. $M_{k,i}$ and $M_{d,i}$ denote the $i$-th element of $M_k$ and $M_d$, respectively. $M_{0,i}$ means the $i$-th element of the geomagnetic vector at the starting point of the navigation. Note that the objective function is normalized to exclude the difference of magnitude between different geomagnetic elements. Under this objective function, the value of $F({M_k})$ will be zero when an AUV arrives at the destination.

Searching a unique solution in \Cref{eq:9} can take a long time. To avoid lengthy searching, the navigation process is terminated when the error converges within a threshold \(\varepsilon \), as shown below
\begin{align}
F({M_k}) \le \varepsilon.
\label{eq:10}   
\end{align}

Above we describe the navigation using geomagnetic information. The performance of this navigation is measured by an objective function, which determines when to terminate the navigation iteration. However, geomagnetic anomalies exist that renders the calculation of declination/inclination gradient in \Cref{eq:6} useless, thus making the calculation of heading angle in \Cref{eq:7} meaningless. Therefore, there is a need to detect and compensate geomagnetic anomalies for a robust geomagnetic navigation. Our study propose a data-driven solution that provides resilient navigation against anomalies, particularly for long-distance underwater navigation, as detailed in \Cref{sec:Method}.

\begin{figure}[htbp]
\centerline{\includegraphics[width=\columnwidth]{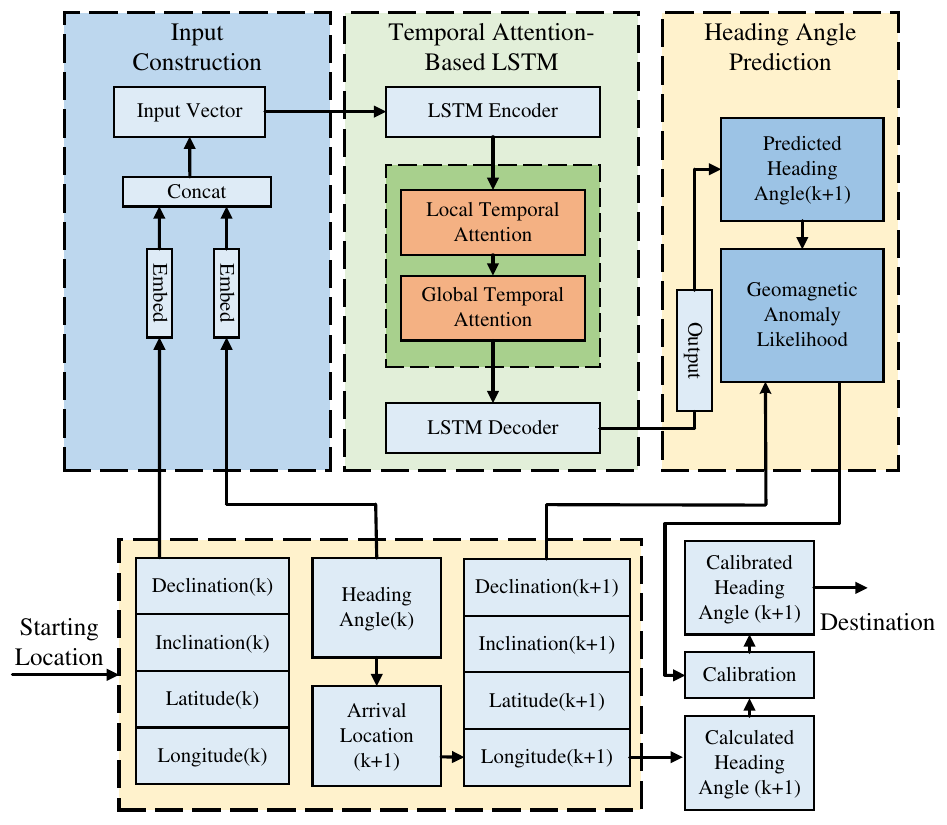}}
\caption{Schematic diagram of our geomagnetic navigation with temporal attention-based Long Short-Term Memory network and anomaly resistance.}
\label{fig:2}
\end{figure}

\section{The Proposed Underwater Navigation with Anomaly Resistance}
\label{sec:Method}
This study comes up with a data-driven approach that predicts the heading angle during the navigation. We achieve this prediction through training a Temporal Attention based Long Short-Term Memory (TA-LSTM) network. The TA-LSTM takes input as time-series of measured geomagnetic information and calculated geographical information, and outputs the heading angle toward the navigation destination. The TA-LSTM penalizes the weights for abnormal elements in an input time-series in case of anomalies, thus providing a robust heading angle prediction. To determine the existence of anomalies, the predicted heading angle by TA-LSTM is compared with the calculated heading angle by \Cref{eq:7}. The final heading angle to be leveraged by the navigation is determined based on the calibration between the predicted and calculated heading angle. A high-level description of the proposed underwater navigation approach is shown in \Cref{fig:2}. 

\begin{figure}[htbp]
\centerline{\includegraphics[width=\columnwidth]{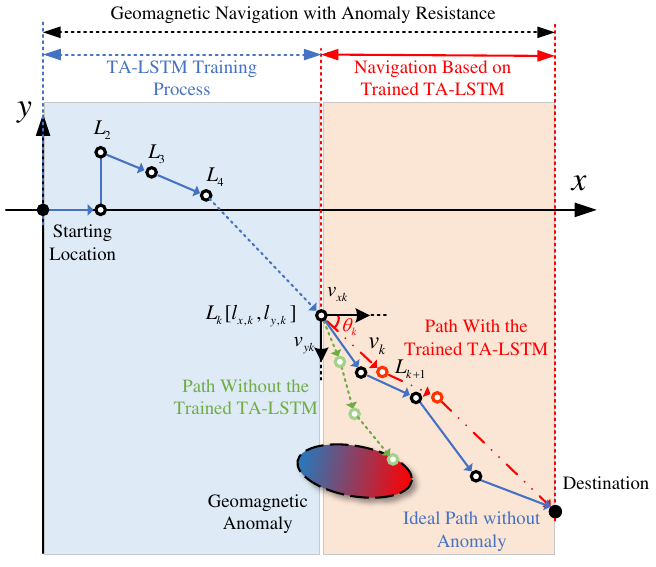}}
\caption{The geomagnetic navigation process of the proposed approach.}
\label{fig:3}
\end{figure}

\Cref{fig:3} demonstrate how the proposed approach with TA-LSTM works during the navigation. The training of the TA-LSTM takes the input of geomagnetic declination, geomagnetic inclination, latitude, and longitude, which are all measured by precise an inertial navigation system (INS)~\cite{Liu2020}~\cite{Xu2017}. The ground-truth for the TA-LSTM output during the traing is also provided by the INS. After the training, the TA-LSTM takes input of the declination/inclination and latitude/longitude calculated from \Cref{eq:6} and \Cref{eq:7}, respectively, to predict the heading angle toward the destination. Below we detail the proposed approach regarding the description of a Long Short-Term Memory (LSTM) network, the encoding and decoding of geomagnetic information in the TA-LSTM, and the anomaly resistance for the navigation with the TA-LSTM.

\subsection{Description of Long Short-Term Memory (LSTM) Networks}
\label{sec:3.1}
An LSTM is a special Recurrent Neural Network (RNN) that addresses the problem of disappearing gradients in RNNs by introducing gating mechanisms~\cite{Hochreiter1997}. With additional memory and gate units, an LSTM is capable of extracting the inherent patterns within a series that exhibit longer intervals and delays~\cite{Zhu2022}. The LSTM consists of hidden state \({h_k}\) and cell memory \({c_k}\), that respectively store the summary of the past input sequences and control the information flow between the input and output through a gating mechanism. The calculation within in an LSTM is as follows:
\begin{align}
{f_k} = \sigma ({W_f} \cdot [{h_{k - 1}},{x_k}] + {b_f}),
\label{eq:11}
\end{align}
\begin{align}
{i_k} = \sigma ({W_i} \cdot [{h_{k - 1}},{x_k}] + {b_i}),
\label{eq:12}
\end{align}
\begin{align}
{o_k} = \sigma ({W_o} \cdot [{h_{k - 1}},{x_k}] + {b_o}),
\label{eq:13}
\end{align}
\begin{align}
{g_k} = \tanh ({W_g} \cdot [{h_{k - 1}},{x_k}] + {b_g}),
\label{eq:14}
\end{align}
\begin{align}
{c_k} = {f_k} \odot {c_{k - 1}} + {i_k} \odot {g_k},
\label{eq:15}
\end{align}
\begin{align}
{h_k} = {o_k} \odot \tanh ({c_k}),
\label{eq:16}
\end{align}
where \({f_k}\) is the output of the forget gate that determines how much information to forget, \({i_k}\) is the output of the gate determines how much information will be used to update the network, and \({o_k}\) is the output of the gate determines how much information will be used for the network ouput. \(\sigma \) is a logistic sigmoid function, and \( \odot \) is element wise product and \({x_k}\) is the input vector. \({W_f}\), \({W_i}\), \({W_o}\) and \({W_g}\) are linear transformation matrices whose parameters need to be learned, while \({b_f}\), \({b_i}\), \({b_o}\) and \({b_g}\) are their corresponding bias vectors. We simplify the LSTM calculations as follows:
\begin{align}
({h_k},{c_k}) = LSTM({x_k},{h_{k - 1}},{c_{k - 1}}).
\label{eq:17}
\end{align}

\subsection{Encoding and Decoding of Geomagnetic Information with Temporal Attention in LSTM}
\label{sec:3.2}
Marine animals can selectively process external information and focus on pertinent signals during long-distance navigation~\cite{Bostroem2012}. We resemble such animal navigation by using Temporal Attention based Long Short-Term Memory (TA-LSTM) network. The temporal attention mechanism in the TA-LSTM can assign higher weights to the most relevant elements in the input time-series. In that way, the heading anlges predicted by the TA-LSTM will be based on reliable geomagnetic information that avoids the impact of geomagnetic anomalies. Below we detail the input and output of the TA-LSTM and how it works in heading angle prediction.

This study uses a multi-input multi-output TA-LSTM, and both the input and output are time-series of length $T$. The $n$-th input time-series is \({X_n} = [x_1^n,x_2^n,...x_k^n,..,x_T^n]\), where
\begin{align}
x_k^n = [L_k^n,S_k^n]^T,
\label{eq:18}
\end{align}
\begin{align}
L_k^n=[{l_{x,((n-1)*T+k)}},{l_{y,((n-1)*T+k)}}],
\label{eq:19}
\end{align}
\begin{align}
S_k^n = [D_{(n-1)*T+k},I_{(n-1)*T+k}],
\label{eq:20}
\end{align}
where $L_k^n$ is the travelled distance at the $((n-1)*T+k)$-th navigation iteration at the x-axis and y-axis, which can be obtained via \Cref{eq:4}. $S_k^n$ denotes the magnetic declination and inclination at the $((n-1)*T+k)$-th navigation iteration, which can be obtained via \Cref{eq:5}.

We describe the $n$-th output of the TA-LSTM as
\begin{align}
Y_n = [y_1^n,y_2^n,...y_k^n,..y_T^n],
\label{eq:21}
\end{align}
where
\begin{align}
y_k^n = \theta_{(n-1)*T+k},
\label{eq:22}
\end{align}
$\theta_{(n-1)*T+k}$ denotes the $((n-1)*T+k)$-th heading angle provided by INS during the TA-LSTM training. In this way, we obtain a collection of input time-series $X = [{X_1},{X_2},..,{X_n},..,{X_N}]$ and the corresponding output time-series $Y = [{Y_1},{Y_2},..,{Y_n},..,{Y_N}]$ that can feed to the training of the TA-LSTM.

\begin{figure}[htbp]
\centerline{\includegraphics[width=0.8\columnwidth]{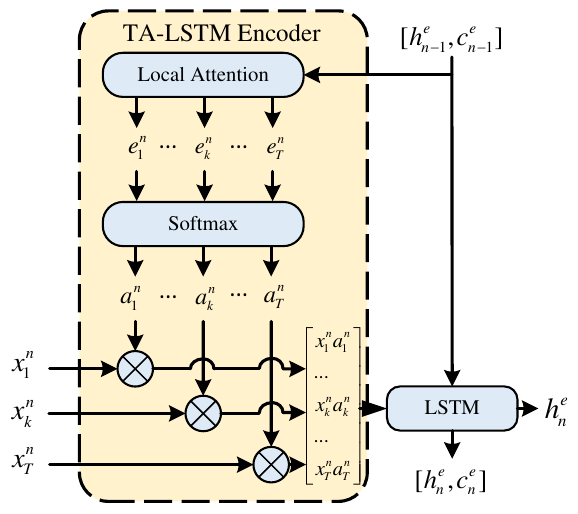}}
\caption{The structure of encoder with local temporal attention.}
\label{fig:4}
\end{figure}

The temporal attention mechanism that differentiate weights for elements in the input time-series is achieved through the encoder-decoder structure in the TA-LSTM. We construct the encoder-decoder structure using two LSTM cells. The encoder is shown in \Cref{fig:4}. %
The encoder takes the $n$-th input as \(x_k^n\) and updates the encoder state vector \(c_n^e\) and hidden vector \(h_n^e\) through equation \Cref{eq:23}. The local attention weight \(e_k^n\), and the weight of importance \(a_k^n\) for the $k$-th element in the $n$-th input time-series are as follows:
\begin{align}
e_k^n = V_e^T \cdot \tanh ({W_e} \cdot [h_{n - 1}^e,c_{n - 1}^e] + {U_e} \cdot {Y_n} + {b_e}),
\label{eq:23}
\end{align}
\begin{align}
a_k^n = {f_{{\rm{softmax}}}}{\rm{(}}e_k^n{\rm{)}},
\label{eq:24}
\end{align}
where \({f_{softmax}}(*)\) is a softmax function. \(V_e^T\), \({W_e}\), \({U_e}\) and \({b_e}\) are the parameters to be learned. 
The hidden state and cell state from the TA-LSTM encoder \(h_n^e\) are as follows:
\begin{align}
(h_n^e,c_n^e) = LSTM({\bar X_n},h_{n - 1}^e,c_{n - 1}^e),
\label{eq:25}
\end{align}
where \({\bar X_n} = (x_1^n \cdot a_1^n,x_2^n \cdot a_2^n,...,x_T^n \cdot a_T^n)\) is the $n$-th input time-series with local attention weight. With the encoder, the information fed into the LSTM is reconstructed and elements in the input are differentiated with different weights.

\begin{figure}[htbp]
\centerline{\includegraphics[width=0.8\columnwidth]{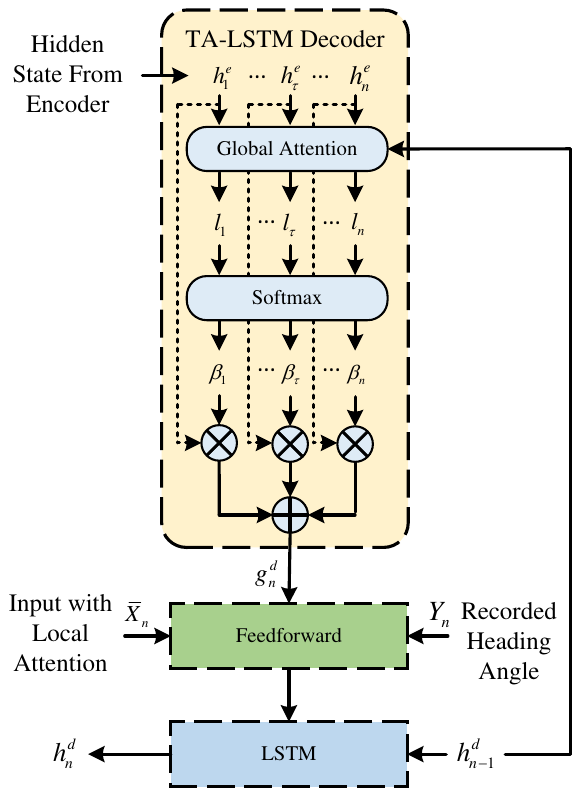}}
\caption{The structure of decoder with global temporal attention.}
\label{fig:5} 
\end{figure}

The TA-LSTM decoder is depicted in \Cref{fig:5}. The decoder uses the learned context from the encoder and makes predictions by attending to specific parts of the input time-series. The decoder takes the hidden states from the encoder in calculating the global temporal attention weights. The global temporal attention weight $g_n^d$ for the $n$-th input time-series is
\begin{align}
{l_n } = V_d^T \cdot \tanh ({W_d} \cdot [h_{n - 1}^d,c_{n - 1}^d] + {U_d} \cdot h_n ^e + {b_d}),
\label{eq:26}
\end{align}
\begin{align}
{\beta _n} = {f_{{\rm{softmax}}}}({l_n}),
\label{eq:27}
\end{align}
\begin{align}
g_n^d = \sum\limits_{\tau= 1}^n {{\beta _\tau }h_\tau ^e},
\label{eq:28}
\end{align}
where \(V_d^T\), \({W_d}\), \({U_d}\) and \({b_d}\) are the parameters to be learned in the global attention calculation, \({\beta _\tau }\) is the global attention weight for the $\tau$-th input time-series. 

The decoder then integrates the $n$-th input time-series with local attention weight $\bar X_n$, the $n$-th global attention weight $g_n^d$, and the $n$-th ground-truth for the output $Y_n$. The integrated information is fed to the input layer of the decoder LSTM, which is a fully connected (FC) layer. The calculation in this FC LSTM layer is as follows
\begin{align}
Y'_n = {W_n} \cdot [{\bar X_n},g_n^d,{Y_n}] + b',
\label{eq:29}
\end{align}
where \({W_n}\) and \(b'\) are the weight and bias for the FC layer, respectively, and $Y'_n$ is the calculation result of the FC layer. The cell state and the hidden state of the decoder LSTM is calculated as
\begin{align}
(c_n^d, h_n^d) = LSTM(Y'_n,h_{n - 1}^d, c_{n-1}^d).
\label{eq:30}
\end{align}

Then we use the $n$-th cell state and the hidden state of the decoder LSTM to predict the output. The $(n+1)$-th predicted output time-series $\overline{Y_{n + 1}} = [\overline{y_1^{n + 1}}, \overline{y_2^{n + 1}}, ..., \overline{y_k^{n + 1}}, ..., \overline{y_T^{n + 1}}]$ is calculated as 
\begin{align}
\overline{Y_{n + 1}} = V_y^T \cdot ({W_y} \cdot [h_T^d,c_T^d] + {b_w}) + {b_y},
\label{eq:31}
\end{align}
where \({W_y}\) and \({b_w}\) are the weight and bias for the output gate in the decoder LSTM, respectively. \(V_y^T\) and \({b_y}\) are the weight and bias of a FC layer for dimension transformation. The predicted heading angle to be used in the next navigation iteration is \(\overline{\theta _k^{n + 1}}=\overline{y_k^{n + 1}}\), $k\in\{1, 2, ..., T\}$. 

The parameters to be learned for the TA-LSTM are identified by minimizing the loss function below
\begin{align}
loss(\phi ) = \frac{1}{M}\sum\limits_{m = 1}^M \left(Y_m - \overline{Y_m}\right)^2,
\label{eq:32}
\end{align}
where, \textit{M} is the number of time-series in the training data set, \(Y_m\) and \(\overline{Y_m}\) are the ground-truth and predicted heading angles for the $m$-th time-series during the training, respectively.

\subsection{Geomagnetic Anomaly and Heading Angle Calibration}
\label{sec:3.3}
This section provides a further step toward navigation resistance against geomagnetic anomalies, along with a robust navigation when the amount of training data is insufficient. Particularly, we employ Maximum Likelihood Estimation (MLE)~\cite{MYUNG200390} to quantify the level of geomagnetic anomaly, and integrate the quantified anomaly level in heading angle prediction, as detailed below. 

We define the error vector for the $n$-th navigation iteration below
\begin{align}
e^{n} = \frac{1}{k}\sum\limits_{k = 1}^T {|{\theta'}_{k}^{n} - \overline{\theta _{k}^{n}}|},
\label{eq:33}
\end{align}%
where \(\theta_{k}^{'n}\) and \(\overline{\theta _{k}^{n}}\) represent the $k$-th element of the $n$-th calculated and predicted heading angle by \Cref{eq:7} and \Cref{eq:31}, respectively. We assume that \(\theta_{k}^{'n} \sim N(\mu ,{\sigma ^2})\) since the geomagnetic signals are distributed in a regular way when there is no geomagnetic anomalies. We use the $n$-th calculated heading angle sequence to estimate the parameters \(\mu \) and \({\sigma ^2}\) via MLE. Then we define an anomaly weight $\eta^{n}$ for the $n$-th calculated heading angle as below
\begin{align}
{\eta^{n}} = \exp ( - \frac{{{{(e^{n} - \mu )}^T}(e^{n} - \mu)}}{{{\sigma ^2}}}).
\label{eq:34}
\end{align}
With the above definition, $\eta^{n}$ decreases when there is a dramatic deviation of geomagnetic signals from expected ones, and $\eta^{n}$ increases when there is no geomagnetic anomalies and the geomagnetic signals distribute normally. We integrate the anomaly weight in heading angle prediction as
\begin{align}
\theta _{k}^{n} = {\eta ^{n}} \cdot \theta _{k}^{'n} + (1 - {\eta ^{n}}) \cdot \bar \theta _k^{n}.
\label{eq:35}
\end{align}
In this way, the prediction of heading angle replies more on the calculated heading angle when the geomagnetic signals distribute normally, otherwise replies more on the heading angle predicted by the TA-LSTM. Thus, the proposed navigation can still work in case of insufficient training data samples yet with normally distributed geomagnetic signals. We summarize our approach in the pseud-code in Algorithm \ref{alg:1}.
\begin{algorithm}
\label{alg:1}
\caption{Geomagnetic Navigation with Anomaly Resistance}
\SetAlgoLined
\KwIn{ \({S_o} = {[{D_o},{I_o}]^T}\), \({S_d} = {[{D_d},{I_d}]^T}\) , \({L_o} = [{l_{x,o}},{l_{y,o}}]\), \({L_d} = [{l_{x,d}},{l_{y,d}}]\), \({\theta _1} = 0\), \({\theta _2} = 90\), \(\varepsilon \), \({V_0}\).}

 Initial input \({X_1}\)  and feature sequence \({Y_1}\) by\Cref{eq:2} -\Cref{eq:7}.

\While{\(F(S_{k + 1}^n) > \varepsilon\)}
{
Calculate heading angle \(\theta _{k}^{'n}\)\;
Calculate local attention weight by\Cref{eq:23} -\Cref{eq:24}\;
Update encoder state \(c_n^e\) and \(h_n^e\) by\Cref{eq:25}\;
Calculate global attention weight by\Cref{eq:26} -\Cref{eq:28} using \Cref{eq:32}\;
Calculate the decoder output by\Cref{eq:29}\;
Update decoder cell and state \(c_n^d\) and \(h_n^d\) by\Cref{eq:30}\;
Predict heading angle \(\overline{\theta _k^{n}}\) by\Cref{eq:31}\;
Finalize the prediction of the heading angle \(\theta _k^{n}\) by\Cref{eq:33} -\Cref{eq:35}\;
Update \(S_{k + 1}^n\) and \(L_{k + 1}^n\) with \(\theta _k^{n}\) and \(V_k^n\) by\Cref{eq:4}.\
 }
 \KwOut{ \(S_{k + 1}^n\), \(L_{k + 1}^n\), \({V_{k}}\), \(\theta_k^{n}\) .}
\end{algorithm}

\section{Simulations and Results}
\label{sec:simulation}
This section carries out simulations to demonstrate the performance of the proposed approach. We conduct the simulations under three navigation conditions: (i) navigation free from geomagnetic anomalies (ii) navigation with geomagnetic anomalies, and (iii) navigation with multiple and consecutive destinations. We employ those diversified conditions to extensively test our approach and make comparisons. Below we explain the specifications that apply to all the three sets of simulations.

We conduct the simulations on a desktop computer (CPU: Intel Core i7-11700: 2.50Ghz; RAM: 16Gb; Graphics: GTX1060: 6Gb) using MATLAB R2023a. We choose a rectangular area for the navigation simulations in the vast western Pacific, which is from 20° north latitude, 130° east longitude (20°N, 130°E) to 30° north latitude, 140° east longitude (30°N, 140°E). According to the WMM 2020~\cite{Chulliat2020}, the geographical coordinates of the area can be uniquely determined by the geomagnetic information. 

We use the WMM2020 model (version of 2020/1/1) to retrieve geomagnetic information. The obtained information of magnetic declination and inclination are transformed to the Universal Transverse Mercator (UTM) coordinate system to empower underwater navigation. We also transform the calculated geographical information to UTM using the \textit{projfwd} function in MATLAB.

We conduct the TA-LSTM network training in the area without geomagnetic anomalies. Geomagnetic navigation requires an appropriate speed of the marine carrier in the beginning as revealed by Zhang \etal~\cite{Zhang2021a}. That is, if the speed is too slow, the geomagnetic gradient cannot be obtained, while a too fast speed leads to unnecessary cost. As such, we chose an initial speed \({V_0}\) as 27 knots (about 50 km/h). To enable a faster convergence in the following navigation process and a nuance speed change near the destination, we set the speed decay for the initial speed as
\begin{align}
\!\!\!V_k^n = \left\{ \begin{array}{l}lr({V_0}),\ if \ |lat{_{K}}\!\! -\!\! lat{_d}| \le {0.5^\circ }\& |lon{_{K}}\!\! -\!\! lon{_d}| \le {0.5^\circ }\\{V_0}, \ otherwise.\end{array} \right.
\label{eq:36}
\end{align}
\begin{align}
K=(n-1)*T+k,
\label{eq:37}
\end{align}
\begin{align}
lr({V_0}) = {\rm{ }}{V_0} * d*{e^{{\rm{int}}(\frac{{T - k}}{{sd}})}},
\label{eq:38}
\end{align}
where, \(lr\) denotes the decay function, \(lat{_{K}}\) and \(lon{_{K}}\) represent the latitude and longitude for the $K$-th location during the navigation, respectively. $int$ means the rounding down to the nearest integer, \textit{sd} denotes the initial speed decay interval, and \textit{d} is the decay rate.

Based on the specifications above, we conduct underwater navigation simulations with diversified navigation conditions, as detailed in the following. 

\subsection{Underwater Navigation without Geomagnetic Anomaly}
\label{sec:4.1}
This simulation uses (22.600°N, 132.900°E) as the starting location and (20.800°N, 136.000°E) as the destination. The geomagnetic information at the starting point is \({S_o}=(D_o, I_0){\rm{ = [ - 4}}{\rm{.321, 30}}{\rm{.861}}{{\rm{]}}^{\rm{T}}}\) and at the destination as \({S_d} =(D_d, I_d)= {[ - 3.425,{\rm{ }}27.008]^T}\). The starting location in UTM is \({L_o} = [{\rm{284116}}{\rm{.65,2500794}}{\rm{.34}}]^T\) and the destination in UTM is \({L_d} = [{\rm{604074}}{\rm{.88,2300365}}{\rm{.07}}]^T\). We implement our approach using the following configurations: $T$=20, $N$=20. The training of the TA-LSTM is configured with the number training epochs as 50, the number of hidden units in the LSTM as 20, the batch size for the input time-series as 20, the learning rate for the training as 0.005, the drop factor in the training as 0.9, the training ratio as 0.7, and the percentage of validation data in the training data as 0.2. We implement another two navigation approaches for comparison: the evolutionary method by Zhang \etal~\cite{Zhang2021} and the 2-D gradient approach by Qi \etal~\cite{DBLP:journals/tgrs/QiXXLR23}, with the same configurations as the work in~\cite{Zhang2021} and~\cite{DBLP:journals/tgrs/QiXXLR23}, respectively. We set the termination condition for the navigation as \(\varepsilon  \leq 0.02\) or a maximum iteration of 300.

\begin{figure}[htbp]
\centerline{\includegraphics[width=0.9\columnwidth]{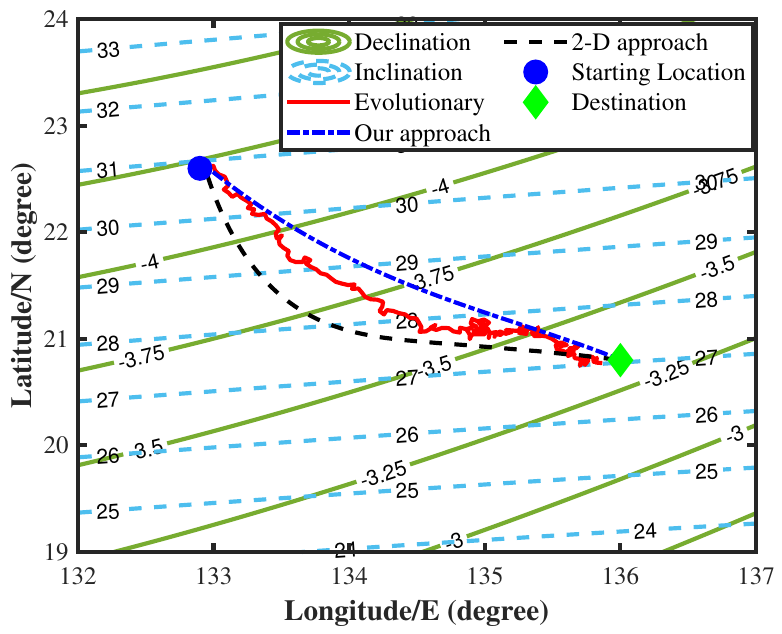}}
\caption{Simulation results without geomagnetic anomalies.}
\label{fig:6}
\end{figure}

We visualize the simulation results in \Cref{fig:6}, where we show the navigation trajectories under the three considered approaches. We can observe a more straight travel path from the origination to the destination by our approach, compared with the evolutionary method and the 2-D approach. We quantify the navigation performance and summarize it in \Cref{tab:1} regarding (i) the final arrival point (ii) the total travelled distance for the navigation (iii) the iterations needed in the navigation (iv) the heading angle variance $\theta _{{\mathop{\rm var}} }$ calculated by\Cref{eq:39}, and (v) the navigation deviation calculated by\Cref{eq:40}. 

\begin{align}
{\theta _{{\mathop{\rm var}} }} = \sum\limits_{n = 1}^N {\sum\limits_{k = 1}^T {(\theta _k^n - \overline{\theta_k^n })/K_t} },
\label{eq:39}
\end{align}

\begin{align}
Deviation = \frac{{{||{L_d} - {L_{K_t}}||} }}{{{||{L_d} - {L_o}||} }},
\label{eq:40}
\end{align}
where $T$ and $N$ denote the length of a single output time-series by the prediction and the total number of output time-series, respectively. $\theta _k^n$ and $\overline{\theta_k^n }$ mean the ground-truth and the predicted heading angle for the $k$-th location of the $n$-th output time-series, respectively. $K_t$ is the total number of locations traversed in the navigation. $L_o$, $L_d$, and $L_{K_t}$ are the origination, destination, and the arrival location for the navigation in UTM coordinate, respectively. \(|| \cdot ||\) denotes the euclidean distance.

\begin{figure}[htbp]
\centering
\subfigure[The Attention-DR convergence curve.]
{
    \begin{minipage}[b]{1\linewidth}
        \centering
        \includegraphics[scale=0.6]{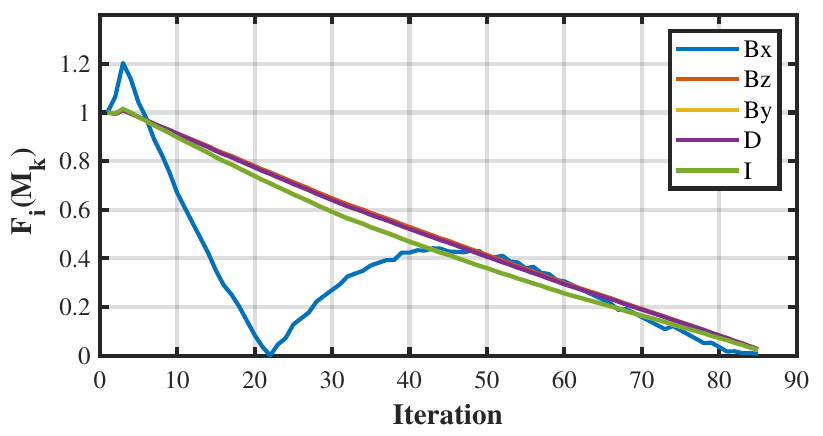}
    \end{minipage}\label{fig:7a}
}
\subfigure[The evolutionary method convergence curve.]
{
 	\begin{minipage}[b]{1\linewidth}
        \centering
        \includegraphics[scale=0.6]{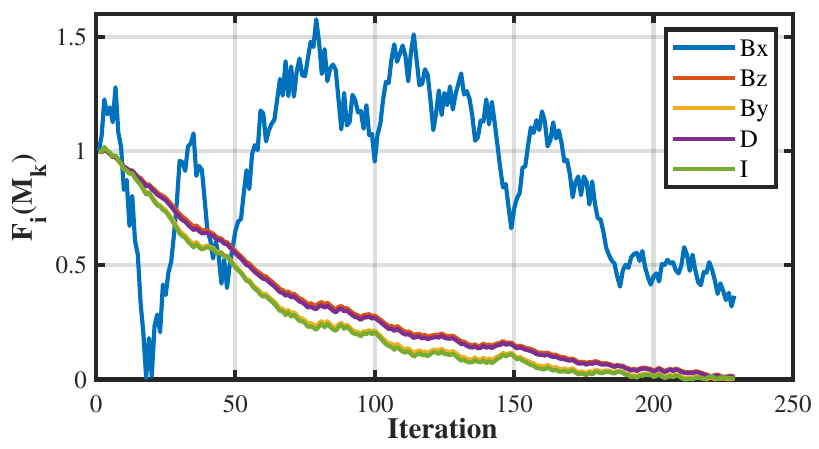}
    \end{minipage}\label{fig:7b}
}
\subfigure[The 2-D approach method convergence curve.]
{
 	\begin{minipage}[b]{1\linewidth}
        \centering
        \includegraphics[scale=0.6]{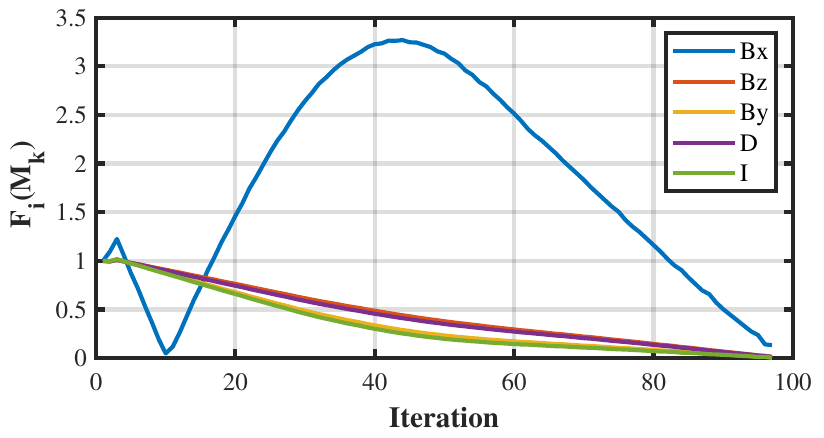}
    \end{minipage}\label{fig:7c}
}
\caption{Convergence of the navigation demonstrated by $F_i(M_k)$, where $i\in\{1, 2, ..., 5\}$, $k$ denotes the iteration number.}
\label{fig:7}
\end{figure} 

In comparison, our approach reduces 62.8\% and 12.3\% navigation iterations compared to the evolutionary method and the 2-D gradient approach, respectively. We also provide a reduction of travelled distance by 46.2\% and 11.1\% compared to the evolutionary method and the 2-D gradient method, respectively. Our approach improves the navigation deviation by 68.2\% and 50.5\% compared to the evolutionary method and the 2-D gradient method, respectively. Based on the results in this simulation, it is evident that the proposed approach provides better navigation performance compared with the evolutionary method and the 2-D gradient method.

\begin{table}[htbp]
    \centering
\caption{Comparison of underwater navigation free from geomagnetic anomalies.}
\label{tab:1}
    \begin{tabularx}{\columnwidth}{cccc} \toprule 
         &  Our approach & Evolutionary & 2-D approach\\ \midrule 
 Arrival location& \makecell[c]{(20.812°N,\\135.954°E)} & \makecell[c]{(20.889°N, \\135.884°E)}& \makecell[c]{(20.862°N,\\135.930°E)}\\ 
         Travel Distance (km)& 380.721& 707.996& 428.495\\ 
 Navigation Steps& 85& 229& 97\\ 
 Heading Angle Variance& 0.1064& 6.1709& 0.2399\\
 Navigation Deviation& 0.0131& 0.0413&0.0265\\
 \bottomrule 
    \end{tabularx}    
\end{table}

\Cref{fig:7} demonstrates the convergence progress of the geomagnetic information from the origination to the destination, measured by \Cref{eq:8}. We use this figure to provide a comprehensive performance illustration and comparison for the navigation. As shown in \Cref{fig:7a}, under the proposed approach, each component of the geomagnetic information except $B_X$ converges to the target at a consistent rate. We observe a similar convergence of geomagnetic information under the evolutionary method and the 2-D approach, shown in \Cref{fig:7b} and \Cref{fig:7c}, respectively, yet with longer navigation iterations. For the geomagnetic information $B_X$, all the three considered approaches cannot provide a expected convergence. However, we observe less fluctuation and lower values of $B_X$ under our approach, compared with the evolutionary method and the 2-D approach. We also observe from \Cref{fig:7b} that the convergence of variables under the evolutionary method is not uniform. This lack of uniform convergence can be a primary factor leading to the rapid fluctuations in heading angle prediction. Though the 2-D method shown in \Cref{fig:7c} provides a smooth geomagnetic information convergence, its dependence on the calculation of geomagnetic gradients without due consideration for positional information can impede and slow down the navigation convergence.

\subsection{Navigation with Geomagnetic Anomalies}
\label{sec:4.2}

\begin{figure}[htbp]
\centerline{\includegraphics[width=0.9\columnwidth]{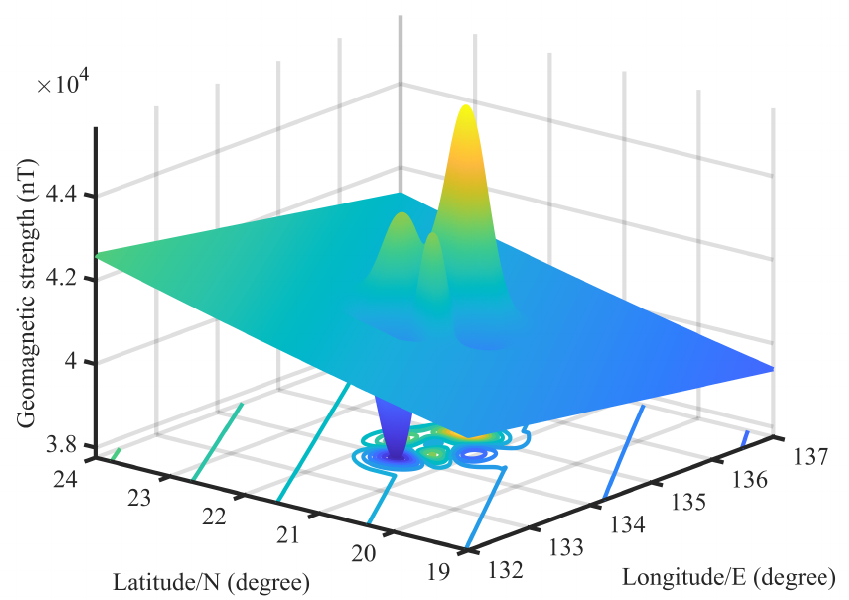}}
\caption{Visualization of the geomagnetic anomaly for the simulation.}
\label{fig:9}
\end{figure}

Geomagnetic anomaly is complex, changeable, and unpredictable, and it is almost inevitable in long-distance navigation. This simulation uses a multimodal function shown in \Cref{eq:41} to generate geomagnetic anomalies. We delineate the area of geomagnetic anomalies within 20°-23°N and 133°-136°E, which is on the anticipated navigation route. The anomaly $B_e$ is mathematically distributed as
\begin{align}
\begin{split}
&{B_e} = 3 \cdot {(1 - x)^2} \cdot {{\mathop{\rm e}\nolimits} ^{\left( { - {x^2} - {{(y + 1)}^2}} \right)}}\\
&- 10 \cdot (\frac{x}{5} - {x^3} - {y^5}) \cdot {{\mathop{\rm e}\nolimits} ^{\left( { - {x^2} - {y^2}} \right)}}- \frac{1}{3} \cdot {{\mathop{\rm e}\nolimits} ^{\left( { - {{(x + 1)}^2} - {y^2}} \right)}},
\end{split}
\label{eq:41}
\end{align}
where $x$, $y$, and $z$ follow the direction of $B_X$, $B_Y$, and $B_Z$, respectively. Through increasing the geomagnetic strength in the $B_X$, $B_Y$, and $B_Z$ by 600, 400, and 200 times, respectively, we obtain the geomagnetic distribution with anomalies as shown in \Cref{fig:9}. Then we implement three navigation methods for comparison: the proposed approach, the evolutionary approach, and the 2-D gradient method, with the same configurations in \Cref{sec:4.1}.

\begin{figure}[htbp]
\centerline{\includegraphics[width=0.9\columnwidth]{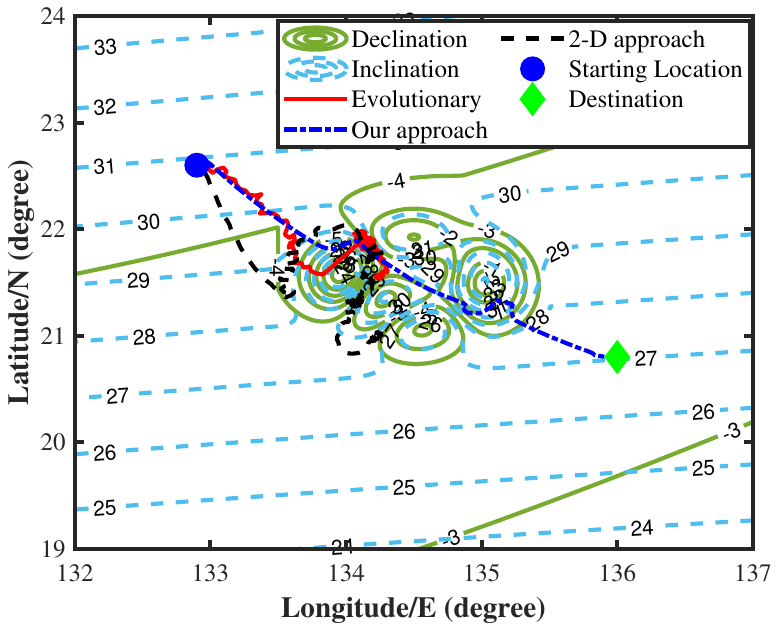}}
\caption{Simulation results under geomagnetic anomaly.}
\label{fig:10}
\end{figure}

\begin{figure}[htbp]
\centering
\subfigure[The Attention-DR convergence curve.]
{
    \begin{minipage}[b]{1\linewidth}
        \centering
        \includegraphics[scale=0.6]{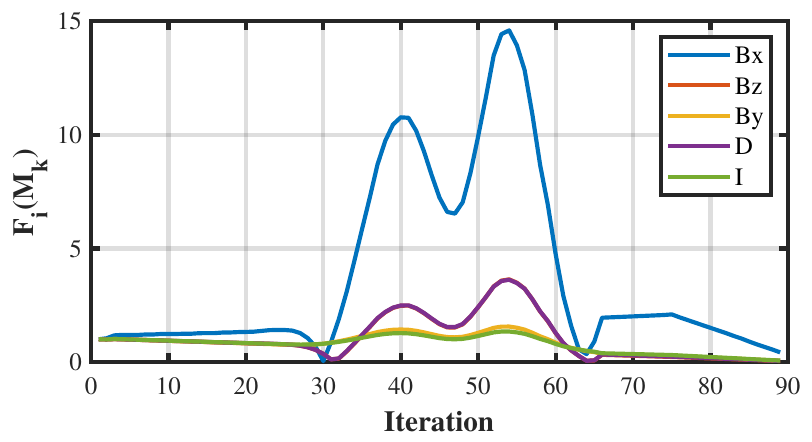}
    \end{minipage}
}
\subfigure[The evolutionary method convergence curve.]
{
 	\begin{minipage}[b]{1\linewidth}
        \centering
        \includegraphics[scale=0.6]{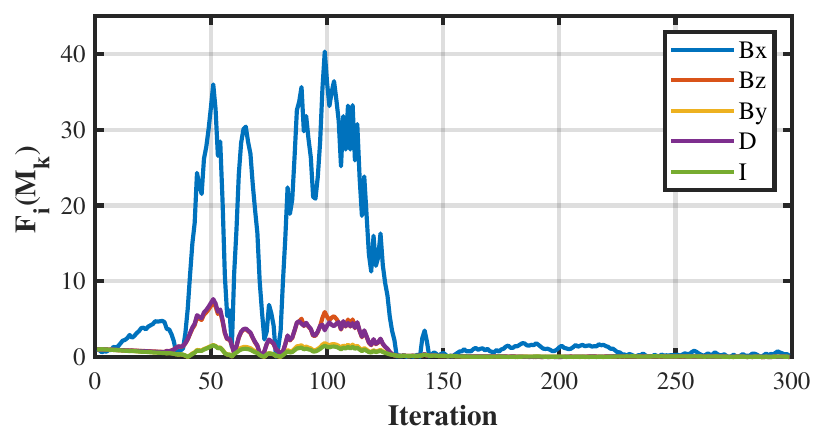}
    \end{minipage}
}
\subfigure[The 2-D approach method convergence curve.]
{
 	\begin{minipage}[b]{1\linewidth}
        \centering
        \includegraphics[scale=0.6]{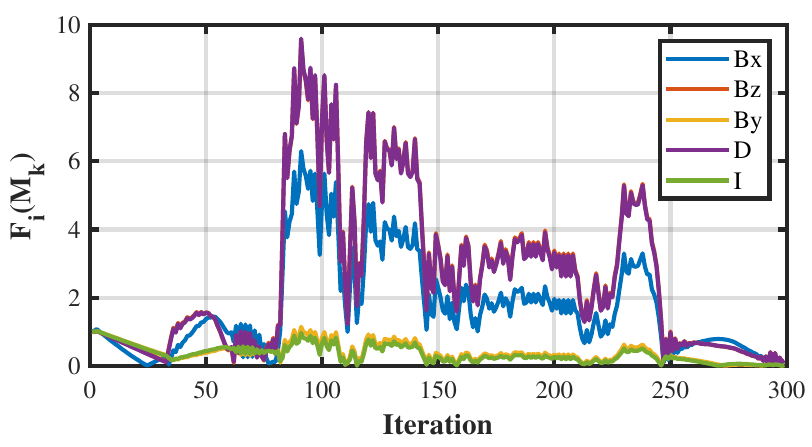}
    \end{minipage}
}
\caption{Convergence of the navigation demonstrated by $F_i(M_k)$, where $i\in\{1, 2, ..., 5\}$, $k$ denotes the iteration number.}
\label{fig:11}
\end{figure}

We show the navigation trajectories under the considered approaches in \Cref{fig:10}, along with the geomagnetic anomalies that disturb the navigation. We can observe that the evolutionary method and the 2-D gradient approach fail to navigate to the destination, with either fluctuations, circles, or distractions from the expected navigation route. The evolutionary method is vulnerable and tends to fall into a local optimal solution, especially in geomagnetic anomaly area, resulting in an inaccuracy calculation of heading angles. The 2-D gradient approach relies on the geomagnetic information of \textit{D} and \textit{I}, yet cannot assess the geomagnetic signal quality and lacks the capability to detect anomalies. As a result, the 2-D gradient approach fails to navigate in geomagnetic anomaly areas with abrupt geomagnetic gradient changes. In the contrast, while the proposed approach also exhibits fluctuations when crossing the geomagnetic anomaly area, it manages to navigate to the destination with a reasonably straight route. \Cref{tab:2} quantifies the navigation performances by the considered approaches regarding the arrival point, travelled distance, navigation steps, heading angle variance, and navigation deviation. 
From \Cref{tab:2}, it is evident that the proposed method accomplishes the navigation mission in the existence of geomagnetic anomalies, whereas the evolutionary method and the 2-D approaching method fail to navigate to the destination under the required iterations. \Cref{fig:11} shows the convergence of geomagnetic signals under the three considered approaches, where we observe more smooth and faster convergence curves by our approach compared with the evolutionary and 2-D gradient method. The evolutionary approach converges faster than the 2-D gradient, yet both of them converge to local optimum and fail to navigate through geomagnetic anomaly area.
\begin{table}[htbp]
    \centering
\caption{Performance comparison of underwater navigation under geomagnetic anomaly.}
\label{tab:2}
    \begin{tabularx}{\columnwidth}{cccc} \toprule 
         &  Attention-DR &  Evolutionary & 2-D approach \\ \midrule 
 Arrival location& \makecell[c]{(20.926°N,\\135.974°E)} & \makecell[c]{(21.249°N,\\135.871°E)}& \makecell[c]{(21.507°N, \\134.194°E)}\\ 
         Travelled Distance (km)& 397.194& 1197.951& 1199.081\\ 
 Navigation Steps& 89& 299& 299\\ 
 Heading Angle Variance& 0.1225& 318.8745& 236.6445\\
 Navigation Deviation& 0.0376& 0.1363&0.5382\\
   \bottomrule 
    \end{tabularx}    
\end{table}

\subsection{Navigation with Multiple Destination}
\label{sec:4.3}
This simulation considers practical navigation where consecutive destinations are required for a single mission. For multiple destination navigation, the heading angles have to vary during the travel, possibly from parallel to orthogonal to the geomagnetic signals. Therefore, calculating the gradient of geomagnetic signals, \eg in the 2-D gradient approach, can be challenging and leads to a nontrivial geomagnetic navigation.

We test our approach for multi-destination navigation in comparison with the 2-D gradient approach. We do not consider the evolutionary method in this simulation. This is because the configurations for the evolutionary method need to change according to each destination, inducing inconsistent experiment settings and complicating the navigation performance comparison.

In this simulation, the navigation commences from Destination $1$ (22.600°N, 132.900°E) to Destination $2$ (22.100°N, 136.00°E), to Destination $3$ (20.3°N, 134.600°E), then to Destination $4$ (19.500°N, 133.500°E), and finally return to Destination $1$. We use the same simulation settings as in \Cref{sec:4.1} and \Cref{sec:4.2}. We relax the termination condition for the arrival of each destination as \(\varepsilon = 0.05\) for a continuous navigation across multiple destinations.
\begin{figure}[htbp]
\centerline{\includegraphics[width=0.9\columnwidth]{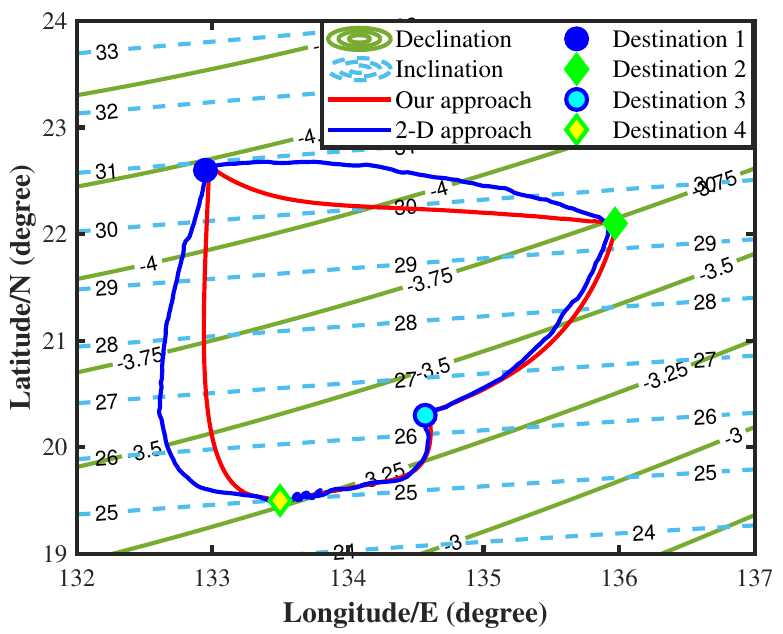}}
\caption{Simulation results for multi-destination navigation.}
\label{fig:12}
\end{figure}

\begin{figure}[htbp]
\centerline{\includegraphics[width=\columnwidth]{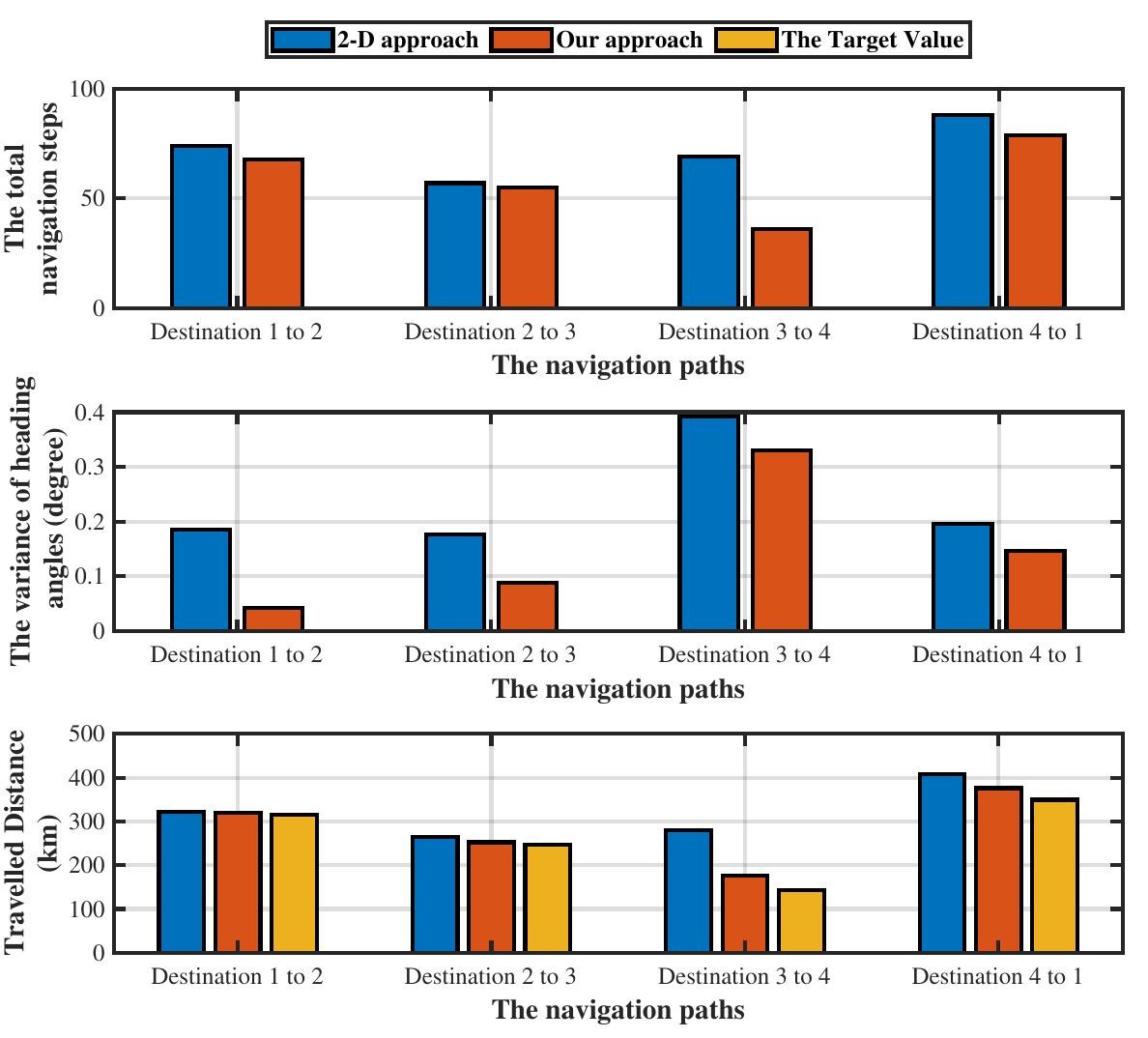}}
\caption{Statistics for the simulation results in multi-destination underwater navigation.}
\label{fig:13}
\end{figure}

We visualize the simulation results in \Cref{fig:12}, where we show the trajectories for the navigation. The figure implies both the proposed and the 2-D gradient method successfully accomplish this multi-destination navigation, while the proposed approach seems to provide more straight travel route. To quantitatively assess the navigation performance, a statistical analysis is performed on 50 simulation results regarding the average navigation iterations, the variance of the heading angle, and the average travelled distance during the navigation. We present the statistical results in \Cref{fig:13}. In this figure, the target value for the travel distance is calculated using precise geographical locations of the origination and the destination. Through the statistics, we observe that the proposed approach provides a 17.4\% reduction in the average navigation iteration compared to the 2-D gradient method, along with a 36.2\% reduction in heading angle variance and a 11.7\% decrease in travelled distance. Note that our approach merely leads to 6.1\% extra travelled distance compared with the target value. In summary, our proposed approach needs less navigation iterations, less distance to cover, and more stable heading angles for multi-distance navigation compared with the 2-D gradient method.

\section{Conclusion}
\label{sec:conclusion}
We proposes a bionic geomagnetic navigation approach for long-distance underwater navigation. Our approach resembles the dead reckoning of marine animals and uses geomagnetic signals for the navigation. We leverage measured geomagnetic data for the navigation, and avoid geographical/geomagnetic maps or related devices. We collect the measured data, and encode the data to construct and train a Temporal Attention based Long Short-Term Memory (TA-LSTM) network for heading angle prediction. Furthermore, we develop the mechanism to detect and quantify geomagnetic anomalies, and calibrate the heading angle prediction by the TA-LSTM. We conduct numerical simulations with diversified navigation conditions to extensively test the proposed approach. The simulation results demonstrate the accurate and stable heading angle prediction for underwater navigation by our approach, and a resilient navigation in geomagnetic anomaly areas.

\bibliographystyle{IEEEtran}
\bibliography{references}

\begin{thebibliography}{10}
\providecommand{\url}[1]{#1}
\csname url@samestyle\endcsname
\providecommand{\newblock}{\relax}
\providecommand{\bibinfo}[2]{#2}
\providecommand{\BIBentrySTDinterwordspacing}{\spaceskip=0pt\relax}
\providecommand{\BIBentryALTinterwordstretchfactor}{4}
\providecommand{\BIBentryALTinterwordspacing}{\spaceskip=\fontdimen2\font plus
\BIBentryALTinterwordstretchfactor\fontdimen3\font minus \fontdimen4\font\relax}
\providecommand{\BIBforeignlanguage}[2]{{%
\expandafter\ifx\csname l@#1\endcsname\relax
\typeout{** WARNING: IEEEtran.bst: No hyphenation pattern has been}%
\typeout{** loaded for the language `#1'. Using the pattern for}%
\typeout{** the default language instead.}%
\else
\language=\csname l@#1\endcsname
\fi
#2}}
\providecommand{\BIBdecl}{\relax}
\BIBdecl

\bibitem{Zhang2021a}
\BIBentryALTinterwordspacing
Y.~Zhang, X.~Liu, M.~Luo, and C.~Yang, ``Bio-inspired approach for long-range underwater navigation using model predictive control,'' \emph{IEEE Trans. Cybern.}, vol.~51, no.~8, pp. 4286--4297, aug 2021. [Online]. Available: \url{https://doi.org/10.1109/tcyb.2019.2933397}
\BIBentrySTDinterwordspacing

\bibitem{MELO2017250}
\BIBentryALTinterwordspacing
J.~Melo and A.~Matos, ``Survey on advances on terrain based navigation for autonomous underwater vehicles,'' \emph{Ocean Engineering}, vol. 139, pp. 250--264, 2017. [Online]. Available: \url{https://www.sciencedirect.com/science/article/pii/S002980181730241X}
\BIBentrySTDinterwordspacing

\bibitem{DBLP:journals/tie/ShenZGCCTLL21}
\BIBentryALTinterwordspacing
C.~Shen, Y.~Zhang, X.~Guo, X.~Chen, H.~Cao, J.~Tang, J.~Li, and J.~Liu, ``Seamless {GPS}/inertial navigation system based on self-learning square-root cubature {Kalman} filter,'' \emph{{IEEE} Trans. Ind. Electron.}, vol.~68, no.~1, pp. 499--508, 2021. [Online]. Available: \url{https://doi.org/10.1109/TIE.2020.2967671}
\BIBentrySTDinterwordspacing

\bibitem{DBLP:journals/ral/ChoiBSC22}
\BIBentryALTinterwordspacing
J.~W. Choi, A.~V. Borkar, A.~C. Singer, and G.~Chowdhary, ``Broadband acoustic communication aided underwater inertial navigation system,'' \emph{{IEEE} Robotics Autom. Lett.}, vol.~7, no.~2, pp. 5198--5205, 2022. [Online]. Available: \url{https://doi.org/10.1109/LRA.2022.3154004}
\BIBentrySTDinterwordspacing

\bibitem{DBLP:journals/tim/WangHWWZ22}
\BIBentryALTinterwordspacing
Z.~Wang, Y.~Huang, M.~Wang, J.~Wu, and Y.~Zhang, ``A computationally efficient outlier-robust cubature kalman filter for underwater gravity matching navigation,'' \emph{{IEEE} Trans. Instrum. Meas.}, vol.~71, pp. 1--18, 2022. [Online]. Available: \url{https://doi.org/10.1109/TIM.2022.3141153}
\BIBentrySTDinterwordspacing

\bibitem{DBLP:journals/inffus/LiZZLNE17}
\BIBentryALTinterwordspacing
Y.~Li, Y.~Zhuang, P.~Zhang, H.~Lan, X.~Niu, and N.~El{-}Sheimy, ``An improved inertial/wifi/magnetic fusion structure for indoor navigation,'' \emph{Inf. Fusion}, vol.~34, pp. 101--119, 2017. [Online]. Available: \url{https://doi.org/10.1016/j.inffus.2016.06.004}
\BIBentrySTDinterwordspacing

\bibitem{DBLP:journals/tgrs/ChenLZLCPHX23}
\BIBentryALTinterwordspacing
Z.~Chen, K.~Liu, Q.~Zhang, Z.~Liu, D.~Chen, M.~Pan, J.~Hu, and Y.~Xu, ``Geomagnetic vector pattern recognition navigation method based on probabilistic neural network,'' \emph{{IEEE} Trans. Geosci. Remote. Sens.}, vol.~61, pp. 1--8, 2023. [Online]. Available: \url{https://doi.org/10.1109/TGRS.2023.3273552}
\BIBentrySTDinterwordspacing

\bibitem{10.1117/12.959127}
\BIBentryALTinterwordspacing
J.~P. Golden, ``Terrain contour matching ({TERCOM}): {A} cruise missile guidance aid,'' in \emph{Image Processing for Missile Guidance}, T.~F. Wiener, Ed., vol. 0238, International Society for Optics and Photonics.\hskip 1em plus 0.5em minus 0.4em\relax SPIE, 1980, pp. 10 -- 18. [Online]. Available: \url{https://doi.org/10.1117/12.959127}
\BIBentrySTDinterwordspacing

\bibitem{7387748}
\BIBentryALTinterwordspacing
B.~Wang, L.~Yu, Z.~Deng, and M.~Fu, ``A particle filter-based matching algorithm with gravity sample vector for underwater gravity aided navigation,'' \emph{IEEE/ASME Transactions on Mechatronics}, vol.~21, no.~3, pp. 1399--1408, 2016. [Online]. Available: \url{https://doi.org/10.1109/TMECH.2016.2519925}
\BIBentrySTDinterwordspacing

\bibitem{DBLP:journals/lgrs/ChenZPCWWL18}
\BIBentryALTinterwordspacing
Z.~Chen, Q.~Zhang, M.~Pan, D.~Chen, C.~Wan, F.~Wu, and Y.~Liu, ``A new geomagnetic matching navigation method based on multidimensional vector elements of {Earth's} magnetic field,'' \emph{{IEEE} Geosci. Remote. Sens. Lett.}, vol.~15, no.~8, pp. 1289--1293, 2018. [Online]. Available: \url{https://doi.org/10.1109/LGRS.2018.2836465}
\BIBentrySTDinterwordspacing

\bibitem{Zhang2021}
\BIBentryALTinterwordspacing
J.~Zhang, T.~Zhang, H.-S. Shin, J.~Wang, and C.~Zhang, ``Geomagnetic gradient-assisted evolutionary algorithm for long-range underwater navigation,'' \emph{IEEE Trans. Instrum. Meas.}, vol.~70, pp. 1--12, 2021. [Online]. Available: \url{https://doi.org/10.1109/tim.2020.3034966}
\BIBentrySTDinterwordspacing

\bibitem{geva2015spatial}
\BIBentryALTinterwordspacing
M.~Geva-Sagiv, L.~Las, Y.~Yovel, and N.~Ulanovsky, ``Spatial cognition in bats and rats: {From} sensory acquisition to multiscale maps and navigation,'' \emph{Nature Reviews Neuroscience}, vol.~16, no.~2, pp. 94--108, 2015. [Online]. Available: \url{https://doi.org/10.1038/nrn3888}
\BIBentrySTDinterwordspacing

\bibitem{DBLP:journals/tgrs/ZhaoHCHLR14}
\BIBentryALTinterwordspacing
Z.~Zhao, T.~Hu, W.~Cui, J.~Huangfu, C.~Li, and L.~Ran, ``Long-distance geomagnetic navigation: Imitations of animal migration based on a new assumption,'' \emph{{IEEE} Trans. Geosci. Remote. Sens.}, vol.~52, no.~10, pp. 6715--6723, 2014. [Online]. Available: \url{https://doi.org/10.1109/TGRS.2014.2301441}
\BIBentrySTDinterwordspacing

\bibitem{putman2020animal}
\BIBentryALTinterwordspacing
N.~F. Putman, ``Animal navigation: {Seabirds} home to a moving magnetic target,'' \emph{Current Biology}, vol.~30, no.~14, pp. R802--R804, 2020. [Online]. Available: \url{https://doi.org/10.1016/j.cub.2020.05.061}
\BIBentrySTDinterwordspacing

\bibitem{gould2014animal}
\BIBentryALTinterwordspacing
J.~L. Gould, ``Animal navigation: {A} map for all seasons,'' \emph{Current Biology}, vol.~24, no.~4, pp. R153--R155, 2014. [Online]. Available: \url{https://doi.org/10.1016/j.cub.2014.01.030}
\BIBentrySTDinterwordspacing

\bibitem{boles2003true}
\BIBentryALTinterwordspacing
L.~C. Boles and K.~J. Lohmann, ``True navigation and magnetic maps in spiny lobsters,'' \emph{Nature}, vol. 421, no. 6918, pp. 60--63, 2003. [Online]. Available: \url{https://doi.org/10.1038/nature01226}
\BIBentrySTDinterwordspacing

\bibitem{naisbett2017magnetic}
\BIBentryALTinterwordspacing
L.~C. Naisbett-Jones, N.~F. Putman, J.~F. Stephenson, S.~Ladak, and K.~A. Young, ``A magnetic map leads juvenile {European} eels to the gulf stream,'' \emph{Current Biology}, vol.~27, no.~8, pp. 1236--1240, 2017. [Online]. Available: \url{https://doi.org/10.1016/j.cub.2017.03.015}
\BIBentrySTDinterwordspacing

\bibitem{fischer2001evidence}
\BIBentryALTinterwordspacing
J.~Fischer, M.~Freake, S.~Borland, and J.~Phillips, ``Evidence for the use of magnetic map information by an amphibian,'' \emph{Animal behaviour}, vol.~62, no.~1, pp. 1--10, 2001. [Online]. Available: \url{https://doi.org/10.1006/anbe.2000.1722}
\BIBentrySTDinterwordspacing

\bibitem{lohmann2004geomagnetic}
\BIBentryALTinterwordspacing
K.~J. Lohmann, C.~M. Lohmann, L.~M. Ehrhart, D.~A. Bagley, and T.~Swing, ``Geomagnetic map used in sea-turtle navigation,'' \emph{Nature}, vol. 428, no. 6986, pp. 909--910, 2004. [Online]. Available: \url{https://doi.org/10.1038/428909a}
\BIBentrySTDinterwordspacing

\bibitem{pakhomov2018magnetic}
\BIBentryALTinterwordspacing
A.~Pakhomov, A.~Anashina, D.~Heyers, D.~Kobylkov, H.~Mouritsen, and N.~Chernetsov, ``Magnetic map navigation in a migratory songbird requires trigeminal input,'' \emph{Scientific Reports}, vol.~8, no.~1, pp. 1--6, 2018. [Online]. Available: \url{https://doi.org/10.1038/s41598-018-30477-8}
\BIBentrySTDinterwordspacing

\bibitem{DBLP:journals/tvt/QiCAWZXYLR18}
\BIBentryALTinterwordspacing
X.~Qi, L.~Chen, K.~An, J.~Wang, B.~Zhang, K.~Xu, D.~Ye, C.~Li, and L.~Ran, ``Bioinspired in-grid navigation and positioning based on an artificially established magnetic gradient,'' \emph{{IEEE} Trans. Veh. Technol.}, vol.~67, no.~11, pp. 10\,583--10\,589, 2018. [Online]. Available: \url{https://doi.org/10.1109/TVT.2018.2866428}
\BIBentrySTDinterwordspacing

\bibitem{taylor2021long}
\BIBentryALTinterwordspacing
B.~K. Taylor, K.~J. Lohmann, L.~T. Havens, C.~M. Lohmann, and J.~Granger, ``Long-distance transequatorial navigation using sequential measurements of magnetic inclination angle,'' \emph{Journal of the Royal Society Interface}, vol.~18, no. 174, p. 20200887, 2021. [Online]. Available: \url{https://doi.org/10.1098/rsif.2020.0887}
\BIBentrySTDinterwordspacing

\bibitem{KIM2023112706}
\BIBentryALTinterwordspacing
Y.~H. Kim, H.~J. Kim, J.~H. Lee, S.~H. Kang, E.~J. Kim, and J.~W. Song, ``Sequential batch fusion magnetic anomaly navigation for a low-cost indoor mobile robot,'' \emph{Measurement}, vol. 213, p. 112706, 2023. [Online]. Available: \url{https://www.sciencedirect.com/science/article/pii/S0263224123002701}
\BIBentrySTDinterwordspacing

\bibitem{DBLP:journals/taes/Canciani22}
\BIBentryALTinterwordspacing
A.~Canciani, ``Magnetic navigation on an {F-16} aircraft using online calibration,'' \emph{{IEEE} Trans. Aerosp. Electron. Syst.}, vol.~58, no.~1, pp. 420--434, 2022. [Online]. Available: \url{https://doi.org/10.1109/TAES.2021.3101567}
\BIBentrySTDinterwordspacing

\bibitem{DBLP:journals/tim/ZhangZSWZ21}
\BIBentryALTinterwordspacing
J.~Zhang, T.~Zhang, H.~Shin, J.~Wang, and C.~Zhang, ``Geomagnetic gradient-assisted evolutionary algorithm for long-range underwater navigation,'' \emph{{IEEE} Trans. Instrum. Meas.}, vol.~70, pp. 1--12, 2021. [Online]. Available: \url{https://doi.org/10.1109/TIM.2020.3034966}
\BIBentrySTDinterwordspacing

\bibitem{DBLP:journals/tcyb/ZhangLLY21}
\BIBentryALTinterwordspacing
Y.~Zhang, X.~Liu, M.~Luo, and C.~Yang, ``Bio-inspired approach for long-range underwater navigation using model predictive control,'' \emph{{IEEE} Trans. Cybern.}, vol.~51, no.~8, pp. 4286--4297, 2021. [Online]. Available: \url{https://doi.org/10.1109/TCYB.2019.2933397}
\BIBentrySTDinterwordspacing

\bibitem{DBLP:journals/tgrs/QiXXLR23}
\BIBentryALTinterwordspacing
X.~Qi, K.~Xu, Z.~Xu, H.~Li, and L.~Ran, ``Geographic true navigation based on real-time measurements of geomagnetic fields,'' \emph{{IEEE} Trans. Geosci. Remote. Sens.}, vol.~61, pp. 1--10, 2023. [Online]. Available: \url{https://doi.org/10.1109/TGRS.2023.3312164}
\BIBentrySTDinterwordspacing

\bibitem{maus2005signature}
\BIBentryALTinterwordspacing
S.~Maus and H.~L{\"u}hr, ``Signature of the quiet-time magnetospheric magnetic field and its electromagnetic induction in the rotating earth,'' \emph{Geophysical Journal International}, vol. 162, no.~3, pp. 755--763, 2005. [Online]. Available: \url{https://doi.org/10.1111/j.1365-246X.2005.02691.x}
\BIBentrySTDinterwordspacing

\bibitem{Glatzmaier1996}
\BIBentryALTinterwordspacing
G.~A. Glatzmaier and P.~H. Roberts, ``Rotation and magnetism of {Earth}{\textquotesingle}s inner core,'' \emph{Sci}, vol. 274, no. 5294, pp. 1887--1891, dec 1996. [Online]. Available: \url{https://doi.org/10.1126/science.274.5294.1887}
\BIBentrySTDinterwordspacing

\bibitem{Komolkin2017}
\BIBentryALTinterwordspacing
A.~V. Komolkin, P.~Kupriyanov, A.~Chudin, J.~Bojarinova, K.~Kavokin, and N.~Chernetsov, ``Theoretically possible spatial accuracy of geomagnetic maps used by migrating animals,'' \emph{J. Roy. Soc. . Interface}, vol.~14, no. 128, p. 20161002, mar 2017. [Online]. Available: \url{https://doi.org/10.1098/rsif.2016.1002}
\BIBentrySTDinterwordspacing

\bibitem{Liu2020}
\BIBentryALTinterwordspacing
B.~Liu, S.~Wei, J.~Lu, J.~Wang, and G.~Su, ``Fast self-alignment technology for hybrid inertial navigation systems based on a new two-position analytic method,'' \emph{IEEE Trans. Ind. Electron.}, vol.~67, no.~4, pp. 3226--3235, apr 2020. [Online]. Available: \url{https://doi.org/10.1109/tie.2019.2910045}
\BIBentrySTDinterwordspacing

\bibitem{Xu2017}
\BIBentryALTinterwordspacing
J.~Xu, H.~He, F.~Qin, and L.~Chang, ``A novel autonomous initial alignment method for strapdown inertial navigation system,'' \emph{IEEE Trans. Instrum. Meas.}, vol.~66, no.~9, pp. 2274--2282, sep 2017. [Online]. Available: \url{https://doi.org/10.1109/tim.2017.2692311}
\BIBentrySTDinterwordspacing

\bibitem{Hochreiter1997}
\BIBentryALTinterwordspacing
S.~Hochreiter and J.~Schmidhuber, ``{Long Short-Term Memory},'' \emph{Neural Comput.}, vol.~9, no.~8, pp. 1735--1780, nov 1997. [Online]. Available: \url{https://doi.org/10.1162/neco.1997.9.8.1735}
\BIBentrySTDinterwordspacing

\bibitem{Zhu2022}
\BIBentryALTinterwordspacing
K.~Zhu, Y.~Li, W.~Mao, F.~Li, and J.~Yan, ``{LSTM} enhanced by dual-attention-based encoder-decoder for daily peak load forecasting,'' \emph{Electr. Pow. Syst. Res.}, vol. 208, p. 107860, jul 2022. [Online]. Available: \url{https://doi.org/10.1016/j.epsr.2022.107860}
\BIBentrySTDinterwordspacing

\bibitem{Bostroem2012}
\BIBentryALTinterwordspacing
J.~E. Boström, S.~{\AA}kesson, and T.~Alerstam, ``Where on earth can animals use a geomagnetic bi-coordinate map for navigation?'' \emph{Ecography}, vol.~35, no.~11, pp. 1039--1047, aug 2012. [Online]. Available: \url{https://doi.org/10.1111/j.1600-0587.2012.07507.x}
\BIBentrySTDinterwordspacing

\bibitem{MYUNG200390}
\BIBentryALTinterwordspacing
I.~J. Myung, ``Tutorial on maximum likelihood estimation,'' \emph{Journal of Mathematical Psychology}, vol.~47, no.~1, pp. 90--100, 2003. [Online]. Available: \url{https://www.sciencedirect.com/science/article/pii/S0022249602000287}
\BIBentrySTDinterwordspacing

\bibitem{Chulliat2020}
\BIBentryALTinterwordspacing
A.~Chulliat, W.~Brown, P.~Alken, C.~Beggan, M.~Nair, G.~Cox, A.~Woods, S.~Macmillan, B.~Meyer, and M.~Paniccia, ``The {US/UK} world magnetic model for 2020-2025: Technical report,'' National Centers for Environmental Information, National Centers for Environmental Information, Tech. Rep., 2020. [Online]. Available: \url{https://doi.org/10.25923/ytk1-yx35}
\BIBentrySTDinterwordspacing

\end{thebibliography}

\begin{IEEEbiography}[{\includegraphics[width=1in,height=1.25in,clip,keepaspectratio]{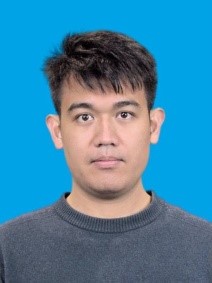}}]{Songnan Yang} received the M.S. degree in Control Science and Engineering in 2020 from Xi’an University of Technology, Xi’an, China. He is currently pursuing his Ph.D. degree in Xi’an University of Technology in Xi’an, China. His current research interests include geomagnetic navigation, machine learning, and signal processing.
\end{IEEEbiography}


\begin{IEEEbiography}[{\includegraphics[width=1in,height=1.25in,clip,keepaspectratio]{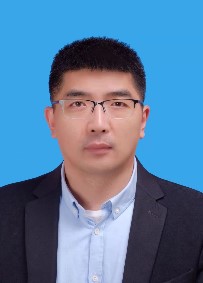}}]{Xiaohui Zhang}
received the B.S. degree in Industrial Automation in 1995, the M.S. degree in Control Science and Engineering in 2002, and the Ph.D. degree in 2009, all from Xi’an University of Technology, Xi’an, China. 

He is currently a Professor at the Department of Information and Control in Xi’an University of Technology, Xi’an, China. His recent research interests include advanced navigation, unmanned systems, signal processing, and pattern recognition.
\end{IEEEbiography}


\begin{IEEEbiography}[{\includegraphics[width=1in,height=1.25in,clip,keepaspectratio]{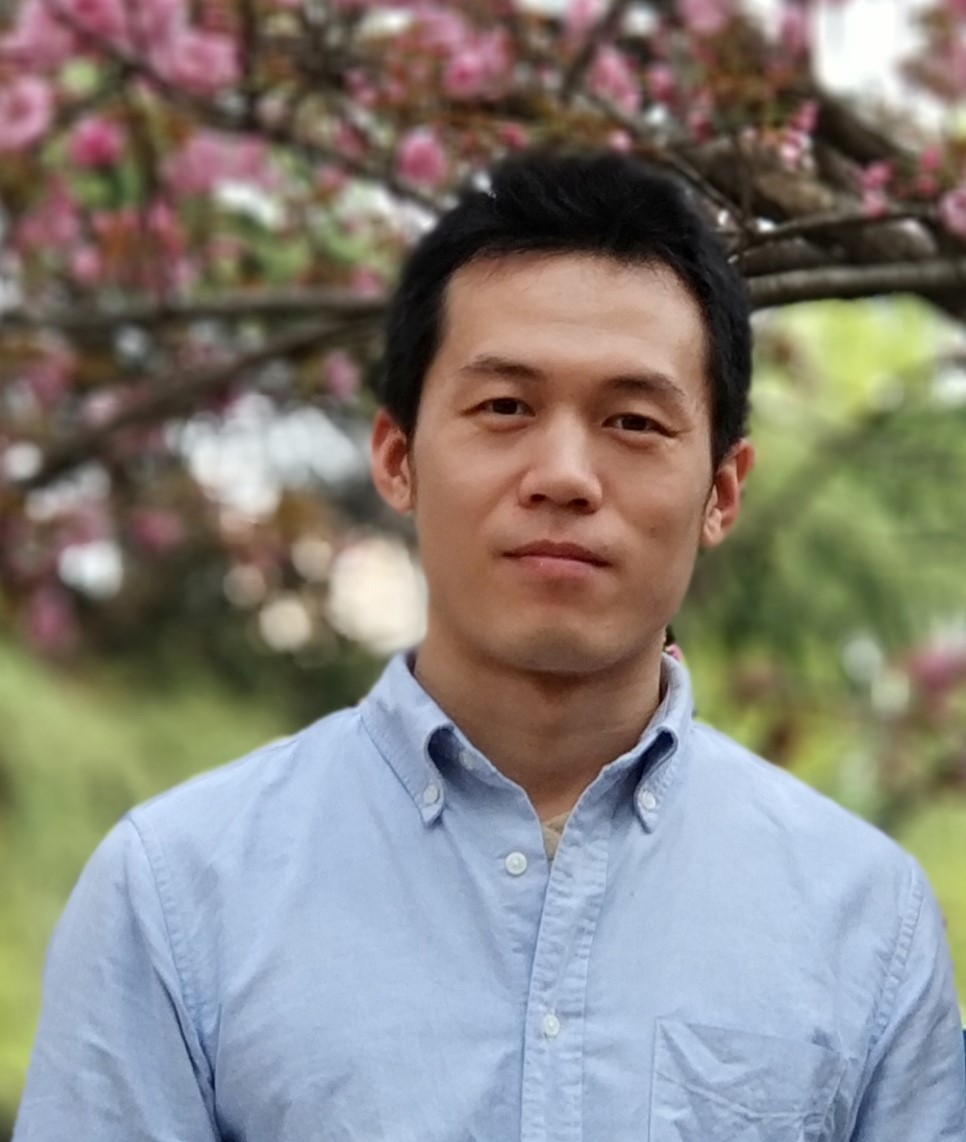}}]{Shiliang Zhang}
(Member, IEEE) received his Ph.D. in electrical engineering from Xi’an Jiaotong University in 2018. He is now a postdoc in the Department of Informatics, University of Oslo, Norway, with his current research focusing on energy informatics, machine/deep learning, system control, and anomaly detection. He previously worked as a postdoc researcher at Chalmers University of Technology in Gothenburg, Sweden, in the field of privacy-preserving machine learning approaches in vehicle networks.
\end{IEEEbiography}


\begin{IEEEbiography}[{\includegraphics[width=1in,height=1.25in,clip,keepaspectratio]{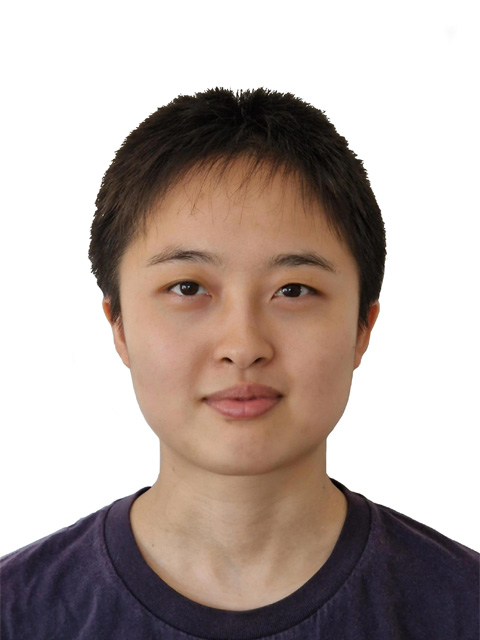}}]{Xuehui Ma} received the Ph.D. degree in control theory \& control engineering, M.S. in control engineering, and B.S. in electrical engineering and automation from the School of Automation and Information Engineering, Xi'an University of Technology, Xi'an, China, in 2014, 2017, and 2022, respectively. Her current research focuses on stochastic control, dual control, robust control and anomaly detection.
\end{IEEEbiography}


\begin{IEEEbiography}[{\includegraphics[width=1in,height=1.25in,clip,keepaspectratio]{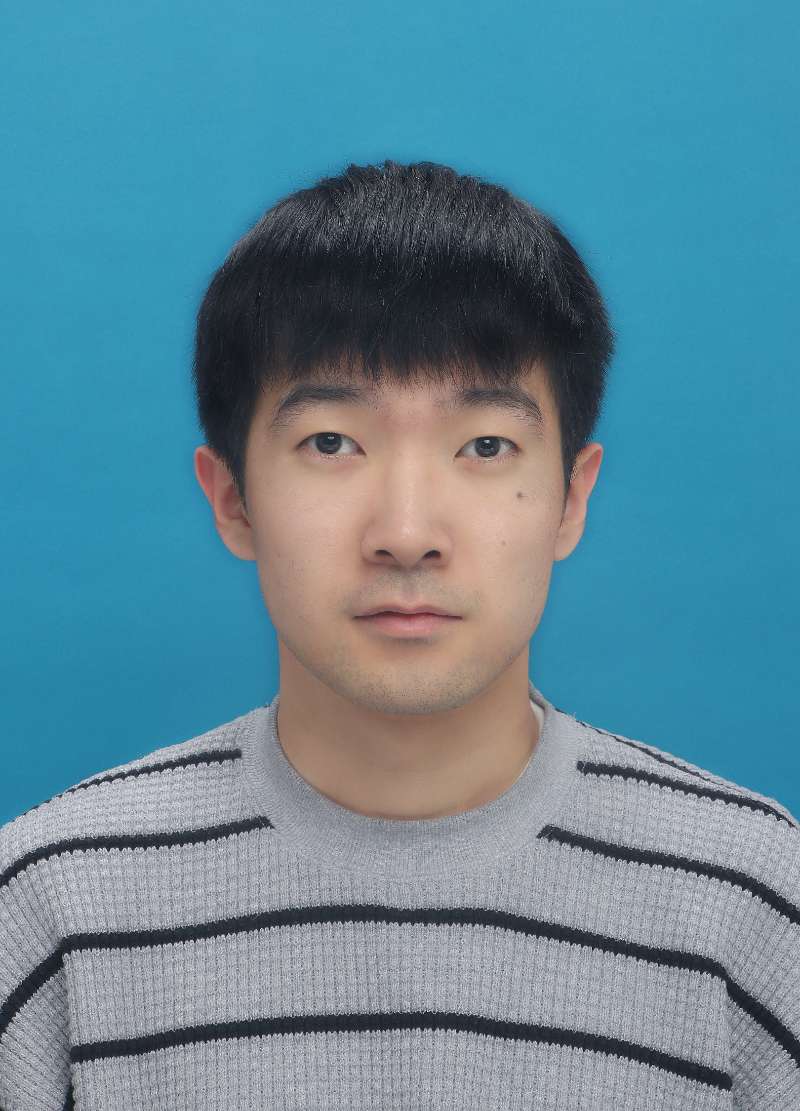}}]{Wenqi Bai}
received the M.S. degree in Control Science and Engineering in 2021 from Xi’an University of Technology, Xi’an, China. He is currently pursuing his Ph.D. degree in Xi’an University of Technology in Xi’an, China. His current research interests include digital twin, reinforcement learning, and geomagnetic navigation.
\end{IEEEbiography}


\begin{IEEEbiography}[{\includegraphics[width=1in,height=1.25in,clip,keepaspectratio]{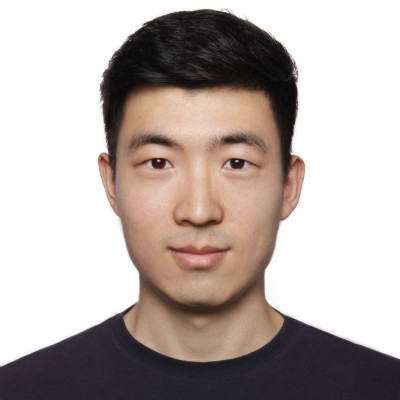}}]{Yushuai Li}
(Member, IEEE) received the B. S. degree in electrical engineering and automation, and the Ph.D. degree in control theory and control engineering from the Northeastern University, Shenyang, China, in 2014 and 2019, respectively. He is currently an Assistant Professor at the Department of Computer Science, Aalborg University, Denmark.

He received the Best Paper Awards from Journal of Modern Power Systems and Clean Energy and 2023 International Conference on Cyber-energy Systems and Intelligent Energies (ICCSIE). His main research interests include distributed optimization and control, machine learning, digital twin, and their applications in integrated energy and transportation systems.
\end{IEEEbiography}


\begin{IEEEbiography}[{\includegraphics[width=1in,height=1.25in,clip,keepaspectratio]{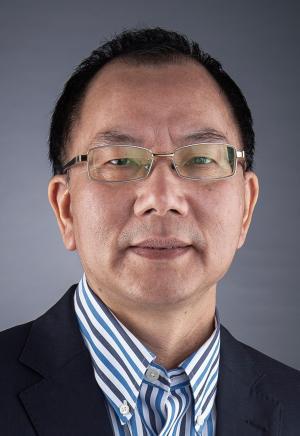}}]{Tingwen Huang}
(Fellow, IEEE) received the B.S. degree in mathematics from Southwest University, Chongqing, China, in 1990, the M.S. degree in applied mathematics from Sichuan University, Chengdu, China, in 1993, and the Ph.D. degree in applied mathematics from Texas A\&M University, College Station, TX, USA, in 2002. 

He worked as a Visiting Assistant Professor with Texas A\&M University. Then, he was an Assistant Professor with Texas A\&M University at Qatar, Doha, Qatar, in August 2003, and then he was promoted to Professor in 2013. His focus areas for research interests include computational intelligence, smart grid, dynamical systems, optimization and control, and cybersecurity.
\end{IEEEbiography}

\end{document}